\documentclass[12pt]{article}

\usepackage{booktabs}    
\usepackage{nicefrac}       
\usepackage{microtype}      
\usepackage{algorithm}
\usepackage{algorithmic}
\usepackage{lipsum}
\usepackage{ragged2e}
\usepackage{geometry}
\usepackage{setspace}

\usepackage[symbol]{footmisc}
\renewcommand*{\thefootnote}{\fnsymbol{footnote}}

\usepackage[title,toc]{appendix}
\usepackage{minitoc}
\noptcrule

\usepackage{natbib}
\usepackage[utf8]{inputenc} 
\usepackage[T1]{fontenc}    
\usepackage{hyperref}   
\hypersetup{
	colorlinks=true,       
	linkcolor=blue,        
	citecolor=blue,        
	filecolor=magenta,     
	urlcolor=blue
}
\usepackage{url}            
\usepackage{booktabs}       
\usepackage{amsfonts}       
\usepackage{nicefrac}       
\usepackage{microtype}      
\usepackage[table,xcdraw]{xcolor}
\usepackage{swstyle}
\usepackage{algorithm}
\usepackage{algorithmic}
\usepackage[]{graphicx,float,latexsym}
\usepackage{amsfonts,amstext,amsmath,amssymb,amsthm}
\usepackage{wrapfig}
\usepackage{lipsum}
\usepackage{gensymb}
\usepackage[normalem]{ulem}

\usepackage{dblfloatfix}

\newcommand{\CommentOFF}{nothinghere}

\usepackage{xcolor}

\newcommand{\todo}[1]{\ifx\CommentOFF\undefined\textcolor{red}{\textsf{TODO: #1}} \fi}

\newcommand{\SFVn}{volume at first $n$-levels in the LOB}
\newcommand{\SFVten}{volume at first $10$-levels}
\newcommand{\SFExecPrf}{cumulative number of executed profile}

\newcommand{\real}{\texttt{real}}

\def\papertitle{INTAGS: Interactive Agent-Guided Simulation}

\title{\papertitle}

\author{Song Wei\textsuperscript{$\dagger$,}\footnote{This work was done while S. Wei was interning at J.P. Morgan AI Research. Contact the authors at: \texttt{song.wei@gatech.edu}, \ \{\texttt{andrea.coletta}, \ \texttt{svitlana.s.vyetrenko}, \ \texttt{tucker.balch}\}\texttt{@jpmchase.com}.}\ , \
    Andrea Coletta\textsuperscript{$\dagger$}, \ 
    Svitlana Vyetrenko\textsuperscript{$\dagger$}, \ and \ Tucker Balch\textsuperscript{$\dagger$} \\
  \small{\textsuperscript{$\dagger$}J.P. Morgan AI Research, \ \textsuperscript{$*$}Georgia Institute of Technology.} 
}

\date{\vspace{-20pt}}

\begin{document}

\maketitle

\begin{abstract} 
In many applications involving multi-agent system (MAS), it is imperative to test an {\it experimental (Exp) autonomous agent} in a high-fidelity simulator prior to its deployment to production, to avoid unexpected losses in the real-world. Such a simulator acts as the environmental {\it background (BG) agent(s)}, called agent-based simulator (ABS), aiming to replicate the complex real MAS. However, developing realistic ABS remains challenging, mainly due to the sequential and dynamic nature of such systems. To fill this gap, we propose a metric to distinguish between real and synthetic multi-agent systems, which is evaluated through the live interaction between the Exp and BG agents to explicitly account for the systems' sequential nature. Specifically, we characterize the system/environment by studying the effect of a sequence of BG agents' responses to the environment state evolution and take such effects' differences as MAS distance metric; The effect estimation is cast as a causal inference problem since the environment evolution is confounded with the previous environment state. Importantly, we propose the \underline{Int}eractive \underline{A}gent-\underline{G}uided \underline{S}imulation (INTAGS) framework to build a realistic ABS by optimizing over this novel metric. To adapt to any environment with interactive sequential decision making agents, INTAGS formulates the simulator as a stochastic policy in reinforcement learning. Moreover, INTAGS utilizes the policy gradient update to bypass differentiating the proposed metric such that it can support non-differentiable operations of multi-agent environments. Through extensive experiments, we demonstrate the effectiveness of INTAGS on an equity stock market simulation example. We show that using INTAGS to calibrate the simulator can generate more realistic market data compared to the state-of-the-art conditional Wasserstein Generative Adversarial Network approach.
\end{abstract}


\newpage

\doparttoc 
\faketableofcontents 

\part{} 


\renewcommand*{\thefootnote}{\arabic{footnote}}

\setstretch{1.3}

\vspace{-0.45in}
\section{Introduction}\label{sec:intro}
In various applications involving multi-agent system (MAS), training or testing an autonomous agent ({\it experimental agent}) that analyzes the current state of the system and responds promptly is a crucial task. Such a task oftentimes relies on live interaction with the environment, be it real or synthetic, to account for the responses of other interactive agents in the environment ({\it background agents}) to the experimental agent's action. 
For example, self-driving algorithms usually analyze large amounts of data from the current state to take decisions, which may condition future states. Thus, they require interactive environments to capture the effect of irregular maneuvers and complex interactions on the other drivers \citep{suo2021trafficsim}, to eventually develop and measure progress. Similarly in the financial domain, algorithmic trading (AT), as a critical component of trading firms, should have access to a live environment with other market participants to construct and test trading strategies \citep{pardo2011evaluation,balch2019evaluate}. 
The live interaction overcomes the limitation of traditional offline methods using historical replays, which fails to capture the dynamic nature of the environment and its reactivity \citep{coletta2023conditional}. 
However, interaction with the real environment is typically expensive, unsafe, and rarely feasible for research purposes, especially in the automotive or financial domain. As a result, the most prevalent approach to develop an experimental agent is through the interaction with a generative model, i.e., {\it agent-based simulator (ABS)}, aiming to emulate the real environment, i.e., background agents' behaviors.

Classic ABS are mostly parametric, and based on explicit rules that the agents need to follow. For instance, under the context of traffic simulation, there is parametric ABS built upon strict rules/laws that the drivers (i.e., agents) must obey \citep{quinlan2010bringing}. However, such rules do not account for irrational or irregular human actions, and they cannot be easily summarized in complex environments, such as financial markets. This complexity motivates the need for advanced machine learning techniques to train non-parametric ABS which can capture those rules from real-world data. 
Focusing on the financial domain, conditional Generative Adversarial Networks (GANs) \citep{goodfellow2014generative} are among the most popular approach for market simulation (see Section~\ref{sec:literature} for related work). Specifically, the state-of-the-art (SOTA) approach is a conditional Wasserstein GAN (cWGAN) trained on market replay, which was recently proposed by \citet{li2020generating,coletta2021towards,coletta2022learning}. To the best of our understanding, the success of cWGAN as SOTA in market simulation comes from its adaption to this specific application: By conditioning on the market input, cWGAN captures historical dependence and attempts to adapt to the sequential nature of trading activities (or rather, the interactions within the MAS).
Additionally, cWGAN uses Wasserstein distance as the discriminator (or metric) to allow synthetic data that is unseen in the historical real market, potentially improving its generalization ability to account for complex market dynamics.

\begin{figure*}[!htp]
\centerline{
\includegraphics[width = \textwidth]{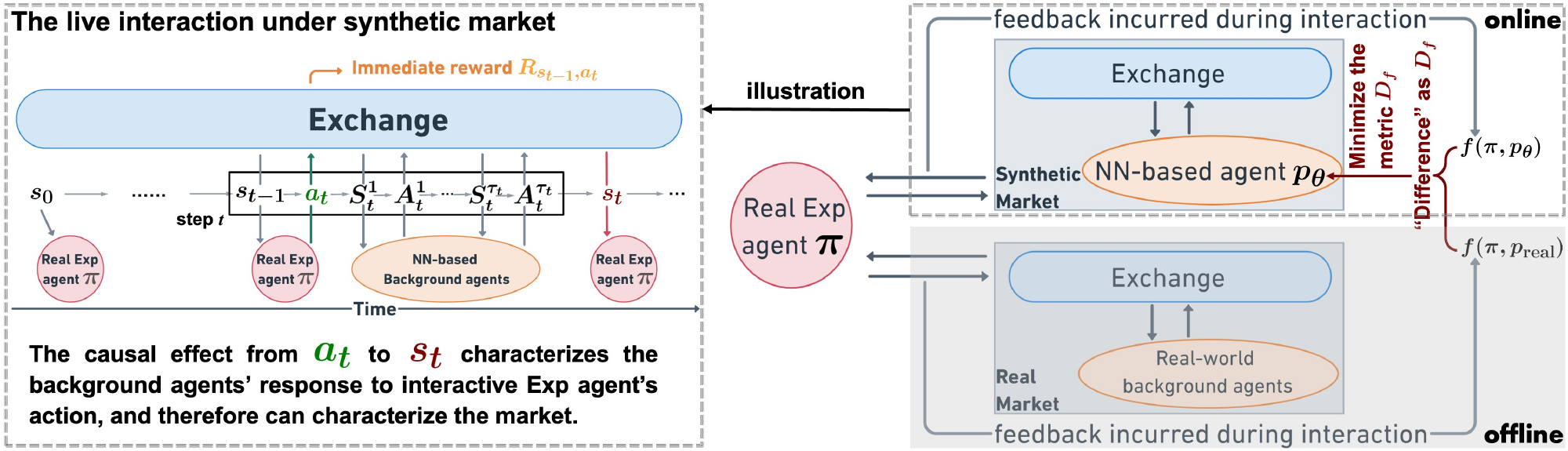}
}
\caption{{Illustration in the market simulation application, where the environment consists of background (BG) agents/traders and the exchange. We show how the experimental (Exp) agent (following policy $\pi$) interact with the BG agents (following policy $p_\theta$ in simulation, or $p_\real$ in reality) in the environment (left) and the INTAGS training framework (right). As in real MAS, within the $t$-th step, the BG agents act $\tau_t$ times, between consecutive actions of the Exp agent.}}
\label{fig:all_illus}
\end{figure*}

However, the aforementioned adaptions are still not able to completely capture the whole market dynamics. {\it First, cWGAN only considers sequential trading activities/interactions locally}. Specifically, cWGAN cuts the sequence of state-action pairs of the background agents into independent pieces. That is to say, cWGAN considers the local dependencies (i.e., the pairs $(S, A)$; please refer to Section~\ref{sec:background} for rigorous definitions of notations) of the whole sequential structure as shown in the left panel of Figure~\ref{fig:all_illus}. Thus, cWGAN-based ABS is merely a generative model that completely ignores the dependency/progression of those sequential pairs.
{\it The second pitfall of cWGAN is its poor generalization ability, which comes from the offline training on replays}: although cWGAN is able to generate (output) unseen states it may not respond realistically when an external agent acts aggressively, creating unseen (input) market states \citep{coletta2023conditional}. During training, cWGAN is conditioned mostly on historical states, with unknown performance when conditioned on states with different data distributions. The cWGAN's response (which shapes the simulated market) to those unseen states may become unrealistic. 
Therefore, it is crucial to introduce external experimental agents and consider the live/online interaction between experimental and background agents during the training.
To the knowledge of the author, even though there are many recent efforts on GAN-based market simulators (to be reviewed in Section~\ref{sec:literature}), they did not go beyond either the classic local metric or offline training on replays, and a satisfying quantitative metric that adapts to the sequential, complex, and dynamic natures of the trading market is largely missing \citep{bouchaud2018trades,vyetrenko2020get}. As a result, it remains a challenging task to train ABS that generates realistic market data.

In this work, we address the aforementioned issues by proposing a distance metric between real and synthetic multi-agent systems that accounts for the sequential and dynamic nature of the MAS. Based on this novel metric, we develop an online training framework, named \underline{Int}eractive \underline{A}gent-\underline{G}uided \underline{S}imulation (INTAGS).
INTAGS only requires offline interactions/replays of the real-world experimental agent with the real environment, and thus will not interfere with the real-world production system; The framework is online because the objective function, or rather the proposed distance metric, is obtained through the online/live interaction of the real-world experimental agent with the synthetic environment.
Specifically, we propose to characterize the underlying environment by the causal effect \citep{rubin1974estimating} from the experimental agent's action to environment state evolution, since such an effect is a result of a sequence of background agents' (which uniquely characterize the environment) responses to the experimental agent's action; We take the difference between the causal effects under real and synthetic environments as our MAS distance metric.
Inspired by SeqGAN \citep{yu2017seqgan}, we formulate the generator network as a stochastic policy to account for the sequential nature of the MAS. 
Moreover, by leveraging the Policy Gradient Theorem \citep{sutton1999policy}, INTAGS minimizes our proposed metric without differentiating the metric, since its evaluation typically involves non-differentiable operations, such as order deletion in our market simulation application. 
We conduct extensive experiments to study the market simulation application, showing our INTAGS can generate much more realistic market data compared with the cWGAN-based simulator.
This improvement could, in turn, facilitate the use of sample inefficient reinforcement learning (RL) approaches for trading strategy construction, reducing ``time-period bias'' and the unrealistic reactivity of the existing cWGAN simulator. 
Although our experiments specifically focus on simulating financial markets, our proposed INTAGS can be applied in all environments with interacting agents, including language generation \citep{bai2022training} and traffic simulation \citep{suo2021trafficsim}.

\subsection{Literature}\label{sec:literature}

One particular application of interest is Limit Order Book (LOB) stock market simulation, a field initially studied via Interactive Agent-Based Simulation \citep{macal2005tutorial} that models and simulates interactions among background agents, i.e., market participants. Presently, the SOTA parametric approach is Agent-Based Interactive Discrete Event Simulation (ABIDES) \citep{byrd2020abides,amrouni2021abides}.
As mentioned earlier, the adaption to the unique characteristics of the underlying application is crucial to the success of generative models. However, oftentimes it would be too difficult to explicitly describe those rules, especially for trading activities, leading to the application of advanced machine learning techniques to summarize those rules from data. 
Under the context of LOB market simulation, precisely specifying the type (momentum agents, value agents, noise agents, etc.) and quantity of background agents to emulate real multi-agent systems is rather challenging, let alone the fact that agents' identities are missing for the MAS calibration \citep{coletta2023k}.
Thus, there is an increasing amount of work replacing parametric models with easy-to-tune neural networks (NNs) to effectively capture the trading market's complex dynamics, offering a more flexible and scalable framework for training a generative model to capture the complex market dynamics. 
Notably, Stock-GAN \citep{li2020generating} utilized cWGAN to generate limit orders;
Later, \citet{coletta2021towards,coletta2022learning} extended Stock-GAN and introduced the concept of a {\it world agent}, which can not only place limit and market orders but also cancel and replace orders.

The aforementioned GAN-based approaches explored various NN architectures, trying to adapt to the corresponding applications, but none of them went beyond the classic local loss that directly penalizes the action of background agents (i.e., the output of the generator) for given input market state --- one exception is \citet{shi2019virtual}, where they studied the customer/agent policy, i.e., how to place BUY orders in the e-commerce market, with RL, and leveraged adversarial imitation learning \citep{ho2016generative,torabi2018generative} that replaced the critic in GAN with the agent reward. 
On one hand, as the generative model tries to capture the policy of the BG agents, it is natural to use RL or imitation learning (IL). In the same spirit of our adaptation of SeqGAN, \citet{shi2019virtual} used IL for generator training to account for the sequential nature of the market. However, to the best of our knowledge, \citet{shi2019virtual} seems to be the only attempt along the direction of IL for market simulation.
On the other hand, as we will see in Section~\ref{sec:exp}, using the reward directly as the loss cannot capture the causal relationship that can discriminate between real and synthetic markets. Additionally, \citet{shi2019virtual} has limitations as it can only place one type of order.

\section{Problem Set-up}\label{sec:background}

We formulate the agents' interactions as a Markov Decision Process, in which the experimental (Exp) agent interacts with the environment (Env), which is characterized by the background (BG) agent(s), to adapt its strategy based on the dynamic feedback from the Env. The Exp agent can be rule-based, or trained by online reinforcement learning, enabling the Exp agent to make informed decisions sequentially in response to changing Env conditions. 

\noindent
\paragraph{Experimental agent.}
Denote the state space by $\tilde{\mathcal{S}}$, action space by $\tilde{\mathcal{A}}$, and a reward function by $R$. At each discrete time step $t \in \{1,\dots,T\}$, where $T$ is a finite time horizon, the Exp agent observes the state $s_{t-1} \in \tilde{\mathcal{S}}$, and takes action $a_t \in \tilde{\mathcal{A}}$ according to its policy $\pi$; this action can incur an immediate reward $R(s_{t-1}, a_t)$ and push the environment to the next state $s_{t}$ as the BG agents respond to $a_t$.

\noindent
\paragraph{Background agent.} 
To distinguish the (synthetic) BG agent from the Exp agent, we denote the input to the BG agent as $S \in \cS$, and output action as $A \in \cA$; Here, $A = G_\theta(z|S), \ z \sim N(\mathbf{0},\mathbf{1})$, where we parameterize $G_\theta$ as a NN with network parameter $\theta$; Indeed, the BG agent follows the underlying stochastic policy:
\begin{equation}\label{eq:worldagent_density}
    A \sim p_\theta(\cdot|S).
\end{equation}
The goal is to train a NN to emulate the real BG agents, ensuring that the simulated environment with a NN-based BG agent closely matches reality. See Figure~\ref{fig:all_illus} for an illustration of the interaction between the Exp agent and BG agent(s). 

\paragraph{Terminologies.} The NN-based BG agent can be viewed as ``one agent representing all real BG agents in the world''; Following idea of the world model \citep{schmidhuber2015learning} and world agent \citep{coletta2021towards}, we name this NN-based BG agent by {\it world BG agent} and the simulated/synthetic interactive environment with world BG agent by {\it world Env}. The interactive environment with real BG agents is referred to as {\it real Env}. In what follows, we will use the terms ``simulator/generator'', ``world BG agent (policy)'', and``world Env'' interchangeably; Similarly, we use the terms ``reality'', ``real BG agents'', and ``real Env'' interchangeably.

\paragraph{Data.}
We use an ordered list $\cT$ to denote the collection of states and actions when the Exp agent $\pi$ interacts with the Env, and we name the collected data during (or after) this interaction as {\it rollout}. Specifically, take the world Env as an example (which is illustrated in Figure~\ref{fig:all_illus}), the {\it partial rollout} that ends at time step $t \in \{1,\dots,T\}$ is: 
\[\cT_{\pi, \theta}^{1:t} = \{s_0, \dots, s_{t-1}, a_t, (S_t^{1},A_t^{1}),\dots,(S_t^{\tau_t},A_t^{\tau_t})\}.\]
In particular, the prefix that determines the last sequential action $A_t^{\tau_t}$ of the BG agent is
\[\Tilde{\cT}_{\pi, \theta}^{1:t} = \left\{s_0, \dots, s_{t-1}, a_t, (S_t^{1},A_t^{1}),\dots,(S_t^{\tau_t-1},A_t^{\tau_t-1}), S_t^{\tau_t}\right\}.\]

\section{Methodology}\label{sec:method}
In this section, we formally define the unique characteristic of the Env, through a  {\it feedback} $f$, after the interaction of a fixed Exp agent $\pi$ with the corresponding Env. Thus, the generator training can be done by minimizing the chosen feedback's difference between the world Env and the real Env.
However, there are two questions: 
\begin{itemize}
    \item \textbf{(Q1)} Which statistics can be used as $f$ to characterize the Env? 
    \item \textbf{(Q2)} Since the evaluation of $f$ after interaction might involve highly non-differentiable operations (such as order deletions from the order book under the context of market simulation), how to minimize the feedback difference between the world and real Envs?
\end{itemize}
Here, we leave $f$ unspecified and answer Q2 by adapting the SeqGAN \citep{yu2017seqgan} to our problem; Later, we will answer Q1 with empirical evidence.

\subsection{Proposed metric between environments}

\paragraph{Feedback $f$.}
For Exp agent with policy $\pi$, the feedback can be obtained through its interaction with the Env, be it real or synthetic. Given the complete rollout $\cT_{\pi, \theta}^{1:T}$ under the world Env, the feedback is denoted by $f(\cT_{\pi, \theta}^{1:T}),$ which can be understood as a realization of random variable $f(\pi, p_\theta)$ shown in Figure~\ref{fig:all_illus}, where the randomness comes from the stochastic policy $p_\theta$ \eqref{eq:worldagent_density}. Similarly, we use subscript ${\real}$ to denote the feedback under real Env, i.e., $f(\cT_{\pi, \real}^{1:T})$, which can be viewed as a realization of random variable $f(\pi, p_{\real})$. 

\paragraph{Market distance metric $D_{f}$.}
Given $f$, our proposed market distance metric, which is highlighted in red in Figure~\ref{fig:all_illus} and will be used to train the world BG agent, is defined as:
\begin{equation}\label{eq:measure}
    \begin{split}
        D_{f}(\theta) = \hat{d}\Big(\big\{ f(\cT_{i,\pi, \theta}^{1:T}), \ i = 1,\dots,N\big\},  \big\{f(\cT_{j, \pi, \real}^{1:T}), \ j = 1,\dots,N'\big\}\Big),
    \end{split}
\end{equation}
where $\cT_{i,\pi, \theta}^{1:T}$'s and $\cT_{j, \pi, \real}^{1:T}$'s are the complete rollouts under the world Env and real Env, respectively.
Here, $\hat{d}$ represents an empirical estimate of a distance metric between probability distributions. A popular example is Maximum Mean Discrepancy (MMD) \citep{gretton2012kernel}. Please see further details, including a graphical illustration of the evaluation of $D_f$ (see Figure~\ref{fig:metric_illus}) and the expression of an unbiased estimate of MMD in Appendix~\ref{appendix:value_eval}.

\subsection{Generator training by minimizing $D_{f}$}
Here, we introduce INTAGS that trains the world BG agent by minimizing interactive agent-based distance metric $D_{f}$ \eqref{eq:measure}. However, the major challenge is that the non-differentiable $D_{f}$ renders commonly seen empirical methods, e.g., back-propagation, for gradient descent (GD) infeasible.
To handle this issue, one popular approach is proposed in SeqGAN work, which reformulated the generator as a stochastic RL policy and performed Policy Gradient update \citep{sutton1999policy}. Fortunately, INTAGS's problem formulation aligns with SeqGAN's in the sense that the world BG agent generates actions sequentially, and the interactive agent-based metric evaluation occurs only at the end of the interaction (to be discussed in Remark~\ref{rmk:endofrollouteval}). Thus, the simulator training is formulated as:
\begin{equation*}
    \theta = \arg\min_{\theta} \cL (\theta) = \sum_{A \in \cA} p_{\theta}(A|S_1^{\tau_1}) Q_{f}(\Tilde{\cT}_{\pi, \theta}^{1:1},A),
\end{equation*}
where the input to world BG agent $S_1^{\tau_1}$ is the last element in the rollout $\Tilde{\cT}_{\theta}^{1:1}$, and $Q_{f}$ is the state-action value function, i.e., the cumulative expected cost conditioned on start state $\Tilde{\cT}_{\theta}^{1:1}$ and action $A$ following the stochastic policy $p_{\theta}$ \eqref{eq:worldagent_density}.

The value function $Q_{f}$ is chosen to be our metric $D_f$ \eqref{eq:measure} with complete rollouts, assuming all intermediate costs are zeros. To be precise, at the intermediate stage of the interaction (say at $t$-th step), $Q_{f}$ can be estimated through $N$ Monte Carlo (MC) simulations to finish the interaction, referred to as {\it MC rollouts}:
\begin{equation*}
    \begin{split}
         Q_{f}(\Tilde{\cT}_{\pi, \theta}^{1:t},A) = \hat{d}\left(f(\boldsymbol{\cT}_{\pi, \theta}^{\rm MC}), f(\boldsymbol{\cT}_{\pi, \real})\right),
    \end{split}
\end{equation*}
where $f(\boldsymbol{\cT}_{\pi, \theta}^{\rm MC})$ denotes the collection of feedbacks obtained from $N$ MC rollouts with prefix $\Tilde{\cT}_{\pi, \theta}^{1:t} \cup \{A\}$, i.e.,
\begin{equation}\label{eq:MCrollout}
    \begin{split}
        f(\boldsymbol{\cT}_{\pi, \theta}^{\rm MC}) =  \Big\{  f\big( \cT_{i,\pi,  \theta}^{1:T, \rm MC}\big):   \Tilde{\cT}_{\pi, \theta}^{1:t} \cup  \{A\}  \subset \cT_{i,\pi, \theta}^{1:T, \rm MC},  \quad i = 1,\dots,N\Big\},
    \end{split}
\end{equation}
and $f(\boldsymbol{\cT}_{\pi, \real})$ denotes the collection of feedbacks obtained from $N^\prime$ complete rollouts under the real Env, i.e., 
\begin{equation}\label{eq:realrollout}
    \begin{split}
        f(\boldsymbol{\cT}_{\pi, \real}) = \left\{f\big(\cT_{j, \pi, \real}^{1:T}\big), \ j = 1,\dots,N^\prime\right\}.
    \end{split}
\end{equation}

The Policy Gradient Theorem gives a closed-form expression of the gradient of the objective function $\cL$ with respect  to (w.r.t.) generator network parameter $\theta$ without differentiating interactive agent-based $Q_f$, i.e.,
\[\nabla_\theta \cL(\theta)=\sum_{t=1}^T \mathbb{E}_{\Tilde{\cT}_{\pi, \theta}^{1:t}}\left[\sum_{A_t \in \cA} \nabla_\theta p_{\theta}\left(A_t |  S_t^{\tau_t}\right) \cdot Q_f\left(\Tilde{\cT}_{\pi, \theta}^{1:t}, A_t\right)\right],\]
where $S_t^{\tau_t}$ is the last element of $\Tilde{\cT}_{\pi, \theta}^{1:t}$. Next,
the likelihood-ratio technique \citep{glynn1990likelihood} is invoked to obtain an unbiased empirical estimate of the gradient, i.e.,
\begin{equation*}
    \begin{split}
 \nabla_\theta \cL(\theta) & \simeq \sum_{t=1}^T \sum_{A_t \in \cA} \nabla_\theta p_\theta\left(A_t |  S_t^{\tau_t}\right) \cdot Q_f\left(\Tilde{\cT}_{\pi, \theta}^{1:t}, A_t\right) \\
& =\sum_{t=1}^T \sum_{A_t \in \cA} p_\theta\left(A_t |  S_t^{\tau_t}\right) \nabla_\theta \log p_\theta\left(A_t |  S_t^{\tau_t}\right) \cdot Q_f\left(\Tilde{\cT}_{\pi, \theta}^{1:t}, A_t\right) \\
& =\sum_{t=1}^T \EE_{A_t \sim p_\theta(\cdot |  S_t^{\tau_t})}\left[\nabla_\theta \log p_\theta\left(A_t |  S_t^{\tau_t}\right) \cdot Q_f\left(\Tilde{\cT}_{\pi, \theta}^{1:t}, A_t\right)\right],
    \end{split}
\end{equation*}
where $\Tilde{\cT}_{\theta}^{1:1} \subset \dots \subset \Tilde{\cT}_{\pi, \theta}^{1:T}$ all come from a complete rollout $\cT_{\pi, \theta}^{1:T}$ under world Env $p_\theta$. The expectation in the last line of the above equation can be approximated via sample average. See Algorithm~\ref{alg:algorithm} for practical implementation of the training of INTAGS. In next section, we will apply INTAGS in a market simulation application, and the complete implementation details can be found in Appendix~\ref{appendeix:method}; For example, one can find a graphical illustration of how to back-propagate to obtain the gradient in Figure~\ref{fig:forward_backward_illus} in Appendix~\ref{appendix:gradient}.

\begin{algorithm}[!htb]
\caption{Training of Interactive Agent-guided Simulation}
\label{alg:algorithm}
\textbf{Input}: Exp agent $\pi$, real feedback $f(\boldsymbol{\cT}_{\pi, \real})$ \eqref{eq:realrollout}, empirical probability distance estimator $\hat d$, learning rate $r$, time horizon $T$, batch size $b$, MC rollout number $N$.
\begin{algorithmic}[1] 
\WHILE{convergence not achieved}
\STATE Perform a complete rollout following $p_\theta$: $\cT_{\pi, \theta}^{1:T}$.
\FOR{$t = 1$ \TO $T$}
\STATE Given prefix $\Tilde{\cT}_{\theta}^{1:t} = \{\dots, \ S_{t}^{\tau_t}\} \subset \cT_{\pi, \theta}^{1:T}$, sample: 
\(A_{t,1},\dots,A_{t,b} \sim p_\theta(\cdot|S_{t}^{\tau_t}).\)
\FOR{$k = 1$ \TO $b$}
\STATE Finish $N$ MC rollouts with prefix $\Tilde{\cT}_{\theta}^{1:t} \cup \{A_{t,k}\}$ following $p_\theta$: $\boldsymbol{\cT}_{\pi, \theta}^{\rm MC}$ \eqref{eq:MCrollout}.
\STATE Compute: $Q_{f}^{t,k} = \hat{d}\left(f(\boldsymbol{\cT}_{\pi, \theta}^{\rm MC}), f(\boldsymbol{\cT}_{\pi, \real})\right).$
\ENDFOR
\ENDFOR
\STATE $\theta \leftarrow \theta - r \sum_{t=1}^T \frac{1}{b} \sum_{k=1}^b Q_{f}^{t,k} \nabla_\theta \log p_\theta\left(A_{t,k} |  S_t^{\tau_t}\right).$
\ENDWHILE
\STATE \textbf{return} $\theta$
\end{algorithmic}
\end{algorithm}

\section{Application to Market Simulation}\label{sec:exp}

In this section, we conduct experiments under the context of financial market simulation to verify the effectiveness of our framework. We will use Env and market interchangeably. 

\subsection{Financial background}\label{sec:financial_background}

\begin{figure}[!htp]
\centerline{
\includegraphics[width = .55\textwidth]{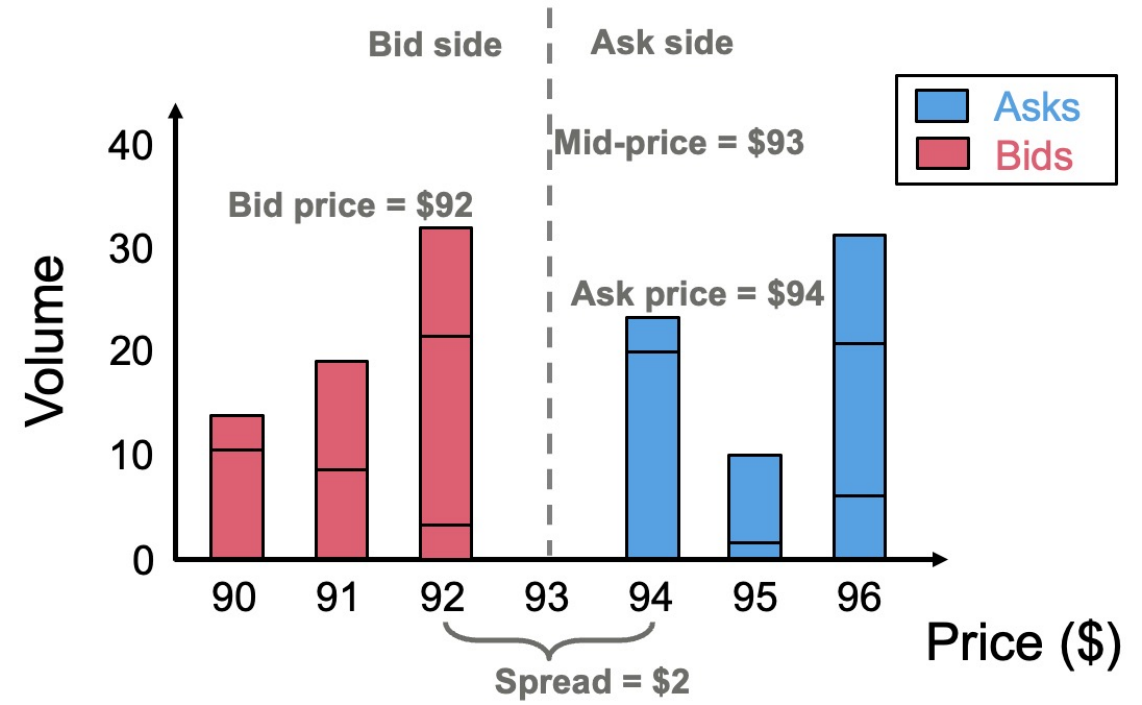}
}
\caption{{The LOB structure for a specific assert.}}
\label{fig:LOB_illus}
\end{figure}

In financial markets, traders buy (bid) and sell (ask) financial assets based on the available market information, aiming to make a profit.
The buy and sell trades/orders are processed through a matching engine, also known as an {\it exchange}, that operates under specific rules (typically ``first in first out'').
The most common orders are market orders and limit orders: Market orders are executed immediately at the prevailing bid or ask price, whereas limit orders set an upper (or lower) price threshold for a buy (or sell) order, and thus may not be (immediately) executed until there is a matching counterparty. 
As illustrated in Figure~\ref{fig:LOB_illus}, the LOB \citep{gould2013limit} organizes and displays outstanding (i.e., unexecuted) limit orders, providing insights into market conditions and liquidity; Its statistical properties, called {\it stylized facts} \citep{cont2001empirical,vyetrenko2020get}, such as mid-price and spread, can help traders with informed decision-making.
See Appendix~\ref{appendix:LOB} for more on LOB and stylized facts. 
One fundamental problem in financial modeling is algorithmic trading (AT), where an algorithmic agent replaces human traders such that it can analyze large volumes of market data, identify patterns, and execute orders at high speeds. Meanwhile, the market responds to the Exp agent's actions in the way that other participants, i.e., BG agents, adjust their strategies and place orders accordingly. This Exp agent can be designed/trained to perform a variety of tasks, e.g., optimal execution, etc. See Appendix~\ref{appendix:opt_exec} for further details.

\subsection{Configurations}\label{sec:exp_config}

In what follows, we use numerical evidence to specify our interactive AT agent-based market distance metric, called {\it AT-based metric}, and demonstrate its effectiveness in training INTAGS. 
Specifically, we name the application of INTAGS under the context of market simulation as Algorithmic Trading-guided Market Simulator (ATMS). 

Since we do not have access to real trading data, we consider a state-of-art parametric simulator as the \textit{real Env}, where it is easier to obtain the ``ground truth''; In particular, we take ABIDES \citep{byrd2020abides} with explicitly specified/parametric BG agents as ``reality''. Using ABIDES data we train our simulated \textit{world BG agent}. The Exp agent is chosen as the AT agent that performs the optimal execution task, aiming to acquire a fixed share of a single asset within a fixed time horizon while minimizing the transaction cost.
To improve simulator resilience under unseen and rare market states \citep{coletta2023conditional}, the AT agent places aggressive orders. Its input includes private state (current time and remaining shares) and market state (such as spread). The world BG agent takes the market state from the past five steps as input \citep{coletta2022learning}. See further details in Appendix~\ref{appendix:training}.

As a disclaimer, neither the method proposed (i.e., our ATMS) nor the AT agent used therein is currently in production. The experiments here are designed to demonstrate the effectiveness of the proposed metric and the training framework, which can support the generalization and deployment of them to reality in the future.

\subsection{Specification of AT-based metric}\label{sec:loss_exp}
\begin{figure}[!htp]
\centerline{
\includegraphics[trim={0.2cm 1.55cm 0.2cm 0.15cm},clip,width = .75\textwidth]{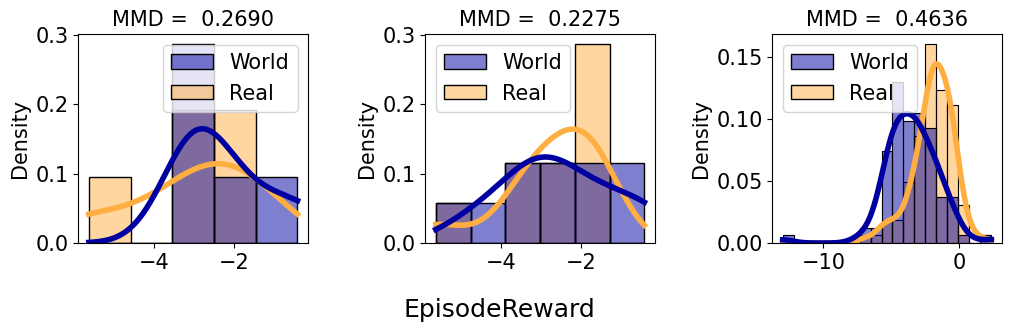}
}
\vspace{0.08in}
\centerline{
\includegraphics[trim={0.2cm 1.55cm 0.2cm 0.15cm},clip,width = .785\textwidth]{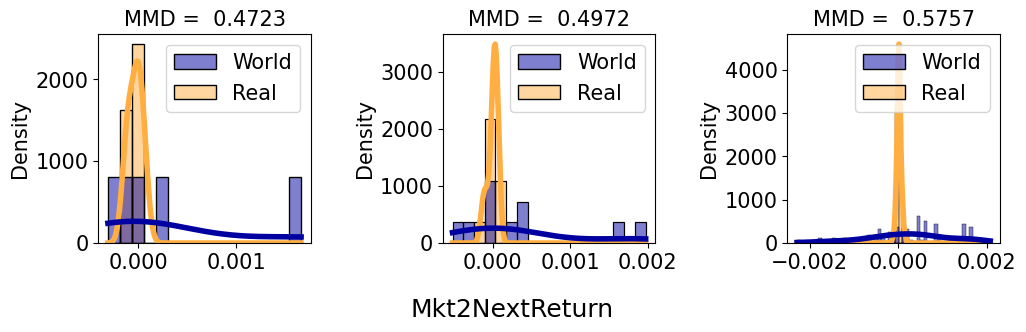}}
\caption{{The distribution of feedback (top: \texttt{Episode} \texttt{Reward}; bottom: \texttt{Mkt2NextReturn}) under multiple rollouts (from left to right: $5, 10, 100$) under real (orange) and world (blue) markets. The ideal feedback should have different empirical distributions under the real and world markets with few rollouts. Thus, \texttt{Mkt2NextReturn}, which can differentiate markets with 5 rollouts as shown in bottom left panel, is better.}}
\label{fig:exp2-1}
\end{figure}

Firstly, we use numerical evidence to select feedback $f$ and empirical probability distance metric $\hat d$.
We start with the most straightforward candidate feedback --- the end-of-rollout cumulative reward of the Exp agent, denoted by \texttt{EpisodeReward} --- and plot its empirical distribution under world and real markets in the first row of Figure~\ref{fig:exp2-1}. 
We can observe that: for a sufficiently large number of rollouts ($N = 100$, the last column in the first row of Figure~\ref{fig:exp2-1}), both visual evidence and quantitative metric MMD support \texttt{EpisodeReward}'s effectiveness in differentiating markets.
However, taking $N = 100$ incurs unreasonably high computational cost --- as shown in Algorithm~\ref{alg:algorithm}, $N$ MC rollouts (step 6) are needed $T \times b$ times for each iteration in ATMS training (see further discussion on complexity at the end of this paper). Given the constraint on $N$, with a limited number of rollouts, it is difficult to differentiate two markets (as shown in the first two columns in the first row of Figure~\ref{fig:exp2-1}), rendering \texttt{EpisodeReward} undesirable for training ATMS.

\paragraph{Axioms for ideal feedback.}
Based on the above analysis, we propose two ``axioms'' that $f$ must satisfy:
\begin{itemize}
    \item \textbf{(Ax1)} Separability --- $f$ must be able to differentiate markets in the sense that $f$'s under different markets should be very different;
    \item \textbf{(Ax2)} Fast convergence --- $f$ should have its distribution quickly ``converge'' to the final state so that only a few rollouts are needed during the ATMS training. 
\end{itemize}
Indeed, \texttt{EpisodeReward} ignores the market dynamics encoded in the time series data collected during the interaction (just as the classic local metric does), and this might explain why the ``convergence'' w.r.t. number of rollouts is not fast enough, resulting in inefficient data usage and information loss.

\paragraph{Causal effect from AT agent action to next-step market return as $f$.} Since the main difference between the real and world markets comes from the BG agents, the feedback should depend on the BG agents.
As illustrated in Figure~\ref{fig:all_illus}, the effect from AT agent action $a_t$ to next state $s_t$ reflects how BG agents respond to $a_t$, and thus the resulting feedback can characterize the BG agents (and thus the market). To be more precise, we consider the action ``placing market order'' since market orders are more frequently seen than limit orders; The stylized fact we consider as the next-step state is market return, which is better (in the sense of the aforementioned 2 axioms) than other candidates including spread, imbalance, etc. For brevity, we name this feedback as \texttt{Mkt2NextReturn}. 


The estimation of \texttt{Mkt2NextReturn} can be cast as a causal/treatment effect estimation problem \citep{rubin1974estimating}, as $s_{t-1}$ acts as a common cause (or rather, confounder) of both action $a_t$ (i.e., treatment) and next state $s_t$ (i.e., observed outcome). To adjust for confounding, we could simply break the dependency between $s_{t-1}$ and $a_t$ by taking a zero-intelligent AT agent, or apply an inverse probability weighted estimator for \texttt{Mkt2NextReturn}; Please see Appendix~\ref{appendix:causal} for further details. 
In the second row of Figure~\ref{fig:exp2-1}, we report the empirical distribution of estimated \texttt{Mkt2NextReturn} against increasing rollout number, from which we can observe that with only $5$ rollouts the empirical distributions are very different from each other (also verified by MMD) and they are fairly close to the ``convergence state'' at $100$ rollouts. Thus, we choose \texttt{Mkt2NextReturn} as $f$ that can differentiate markets with a small number of rollouts. 

For completeness, we report the results for other stylized facts as the next state in Figure~\ref{fig:exp2-4}, which generally performs poorly compared to market return in terms of (Ax1) and (Ax2), suggesting that those stylized facts might be less representative of the market. We want to particularly mention that  using price impact as the next state performs nearly as well as using market return; In the next subsection, we will demonstrate that employing \texttt{Mkt2NextPriceImpact} as $f$ leads to a closer resemblance of \SFExecPrf \ to reality, while \texttt{Mkt2NextReturn} improves the match to reality in terms of \SFVn. Due to space consideration, those results are deferred to Appendix~\ref{appendix:f}.

\begin{figure}[!htp]
\centerline{
\includegraphics[trim={0.2cm 1.55cm 0.2cm 0.15cm},clip,width = .75\textwidth]{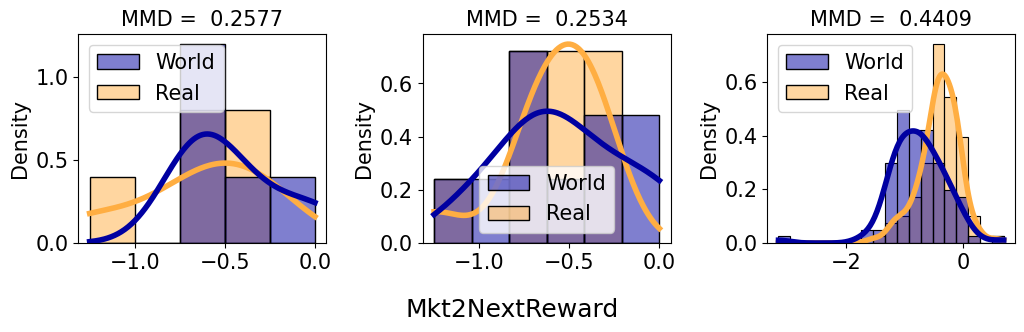}
}
\caption{The distribution of \texttt{Mkt2Reward} when performing multiple (from left to right: $5, 10, 100$) rollouts using AT agent under real (orange) or world BG agent (blue) markets. Apparently, when there are only around 5 rollouts, \texttt{Mkt2Reward} as $f$ cannot faithfully differentiate the markets, making it a poor candidate. This reaffirms our claim the the feedback must depend on the underlying BG agents.}\label{fig:exp2-3}
\end{figure}

In the last experiment to support our feedback choice above, we consider the effect from action $a_t$ to immediate reward $R(s_{t-1},a_t)$, denoted by \texttt{Mkt2Reward}. 
We report the result for \texttt{Mkt2Reward} in Figure~\ref{fig:exp2-3}, which exhibits similar patterns to \texttt{EpisodeReward}, providing strong evidence to support our claim that \texttt{Mkt2Reward} solely depends on the current LOB status and the exchange rule, and thus cannot represent the BG agents.

\begin{figure}[!htp]
\centerline{
\includegraphics[width = .35\textwidth]{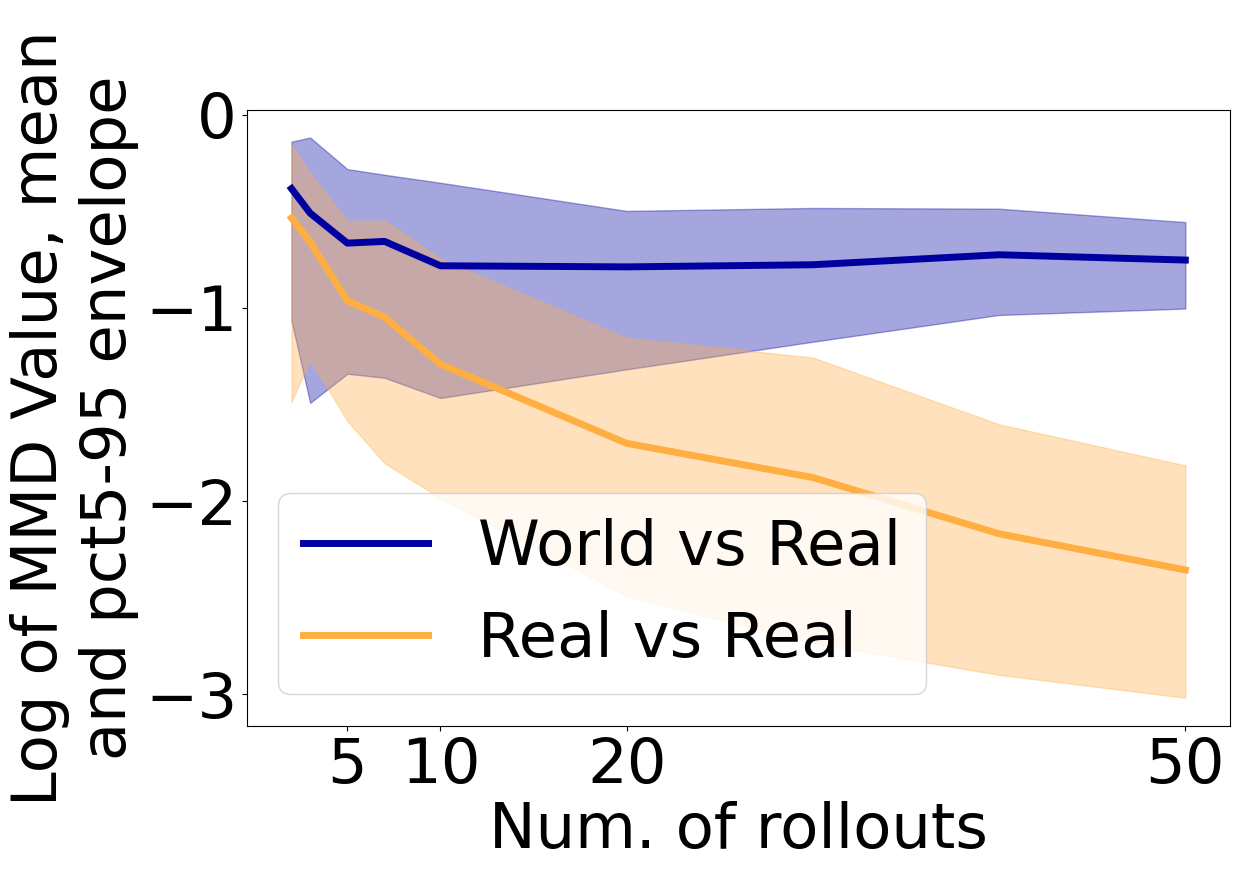}
\hspace{0.1in}
\includegraphics[width = .35\textwidth]{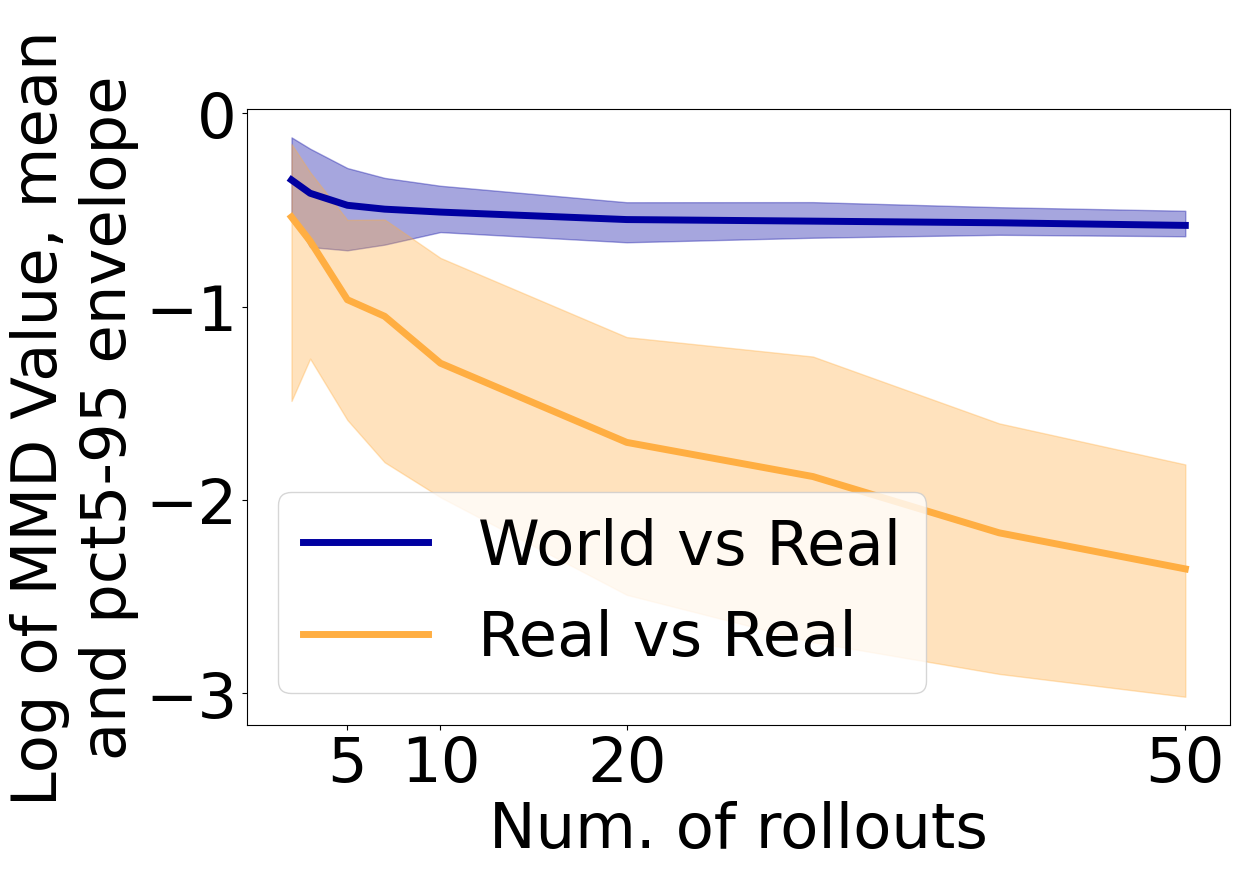}
}
\caption{{The mean and $90\%$ CI trajectory of our metric: $\hat d$ is MMD estimator, and $f$ is \texttt{EpisodeReward} (left) or \texttt{Mkt2NextReturn} (right). The ideal metric should be large (or small) under different (or same) markets with few rollouts, reaffirming our finding from Figure~\ref{fig:exp2-1} that \texttt{Mkt2NextReturn} is a better option.}}
\label{fig:exp3-1} 
\end{figure}

\noindent
\paragraph{MMD as $\hat d$.}
To train ATMS by minimizing 
$D_{f}$ \eqref{eq:measure}, $\hat d$ should also be properly chosen such that $D_{f}$ meets (Ax1) and (Ax2). To ensure that the good separation shown in the second row of Figure~\ref{fig:exp2-1} is not a result of certain random seeds, we perform bootstrap to quantify the uncertainty: For rollout number $N \in \{2,3,5,7,10,20,30,40,50\}$, we select $N$ rollouts from $200$ rollouts and repeat this procedure $50$ times to obtain $50$ MMDs; We plot the mean and $5 \%$ - $95 \%$ envelope, i.e., $90\%$ bootstrap confidence interval (CI), trajectory of those MMDs against $N$ when the two markets are different (or identical) in Figure~\ref{fig:exp3-1}. 
Additionally, we report result for \texttt{EpisodeReward} as $f$ for comparison. We can observe that MMDs under the same market (red) are significantly smaller than that under different markets (blue) when $N$ exceeds $5$, verifying $D_f$'s effectiveness when choosing \texttt{Mkt2NextReturn} as $f$ and MMD as $\hat d$ to differentiate markets. On the contrary, such a pattern can only be observed for large enough $N$ (around $30$) when we consider \texttt{EpisodeReward} as $f$.
For completeness, we report the results for ED and EMD in Figure~\ref{fig:exp3-2} and similar results for other feedback candidates such as \texttt{Mkt2NextReward} in Figure~\ref{fig:exp3-3}, and those results reaffirm our choice that \texttt{Mkt2NextReturn} as $f$ and MMD as $\hat d$ to train the ATMS; Please find those results in Appendix~\ref{appendix:hatd}.

\begin{remark}[Evaluation at the end of interaction]\label{rmk:endofrollouteval}
    Evaluating $f$ as well as $D_f$ using partial rollout (i.e., ignoring step 6 in Algorithm~\ref{alg:algorithm}) is feasible, but will lead to insufficient data for the feedback estimation and metric evaluation, resulting in a less favorable metric that cannot differentiate markets.
\end{remark}

\subsection{Performance improvement of ATMS}\label{sec:finetune_exp}

\begin{figure*}[h]
\centerline{
\includegraphics[width = \textwidth]{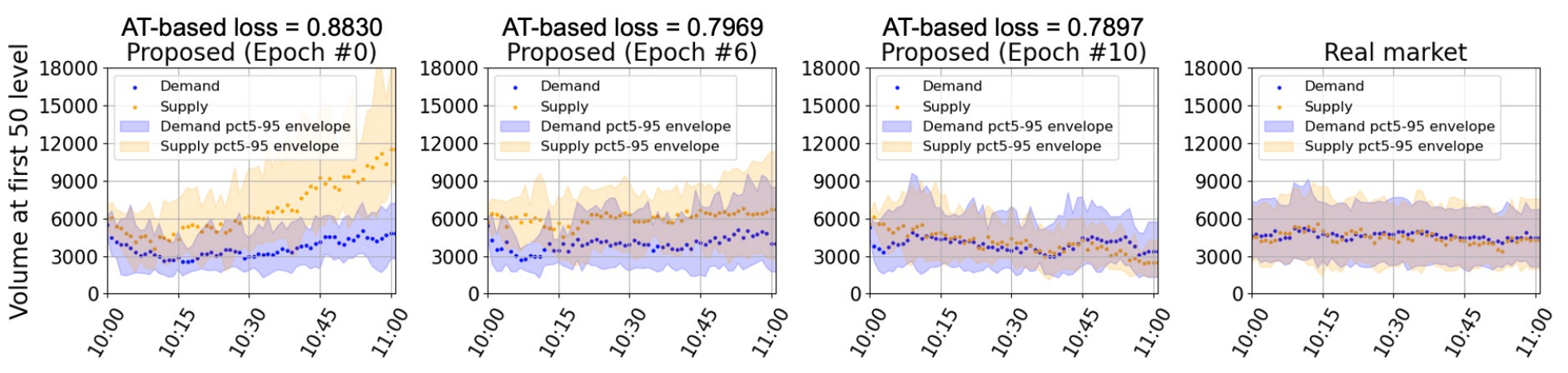}
}
\caption{{Effectiveness of ATMS, which is the application of our INTAGS in finance. We report \SFVten. The ATMS can generate market data that is ``increasingly'' similar to reality while minimizing our AT-based metric.}}
\label{fig:exp_effectiveness_1}
\end{figure*}

We evaluate the simulator performance with not only our novel metric but also visualizing certain stylized fact time series \cite{bouchaud2018trades,li2020generating}. To demonstrate the effectiveness of our ATMS, we visualize the stylized fact \SFVten \ from 10AM to 11AM over 12 independent trials in Figure~\ref{fig:exp_effectiveness_1}. We can observe that, during the ATMS training, {\SFVten} is ``increasingly similar'' (in terms of magnitude and matching demand-supply) to reality while our AT-based metric is minimized. Please see Appendix~\ref{appendix:effectiveness} for further evidence of the effectiveness.

Next, we show that our ATMS outperforms existing cWGAN baseline \citep{coletta2022learning}; See Appendix~\ref{appendix:GAN} for further details. Interestingly, we find that the stylized facts are not only similar to the target market but also show more balanced BUY and SELL volumes. While the cWGAN training can be biased towards one direction \citet{coletta2023conditional}, our novel approach seems to provide a more fair metric and avoid such biases. Thus we could mitigate the weakness shown in \citet{coletta2023conditional}, where a fairly simple, yet effective, strategy can exploit such biases, and turn the BUY/SELL unbalance into a profit.
Indeed, as illustrated in Figure~\ref{fig:all_illus}, 
unlike classic local loss that directly penalizes the world BG agent's actions $A_t^{(j)}$'s for given inputs $S_t^{(j)}$'s, our proposed metric penalizes the AT agent input $s_t$ for given previous step's AT agent action $a_t$. As a result, by considering that AT agent input (or state) $s$ contains numerous stylized facts characterizing the market, ATMS that minimizes our metric can produce stylized facts much closer to reality compared to the cWGAN baseline.

\begin{figure*}[!htp]
\centerline{
\includegraphics[trim={0.25cm 0.25cm 0.25cm 0.25cm},clip,width = .208\textwidth]{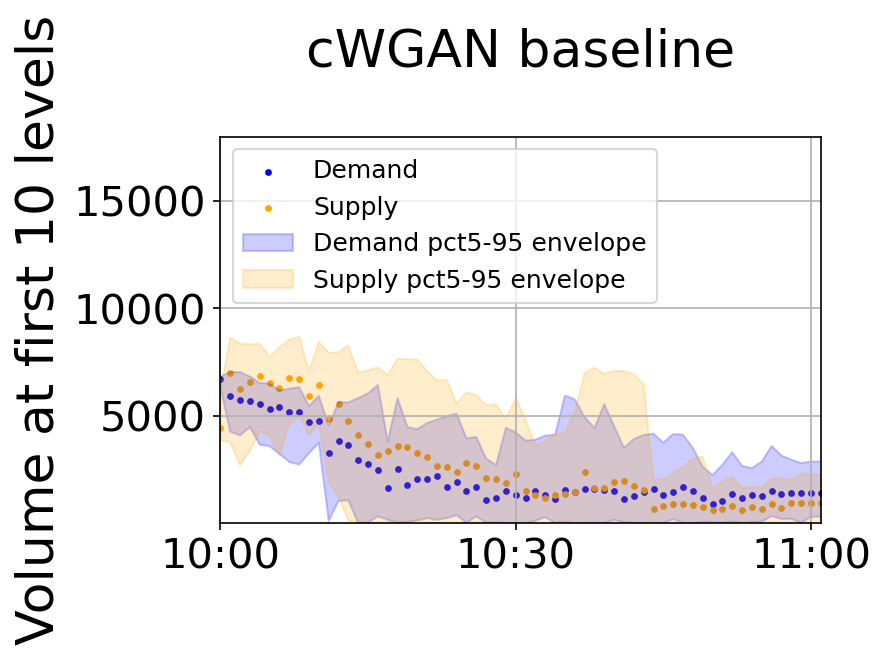}
\hspace{-0.06in}
\includegraphics[trim={1.3cm 0.25cm 0.25cm 0.25cm},clip,width = .195\textwidth]{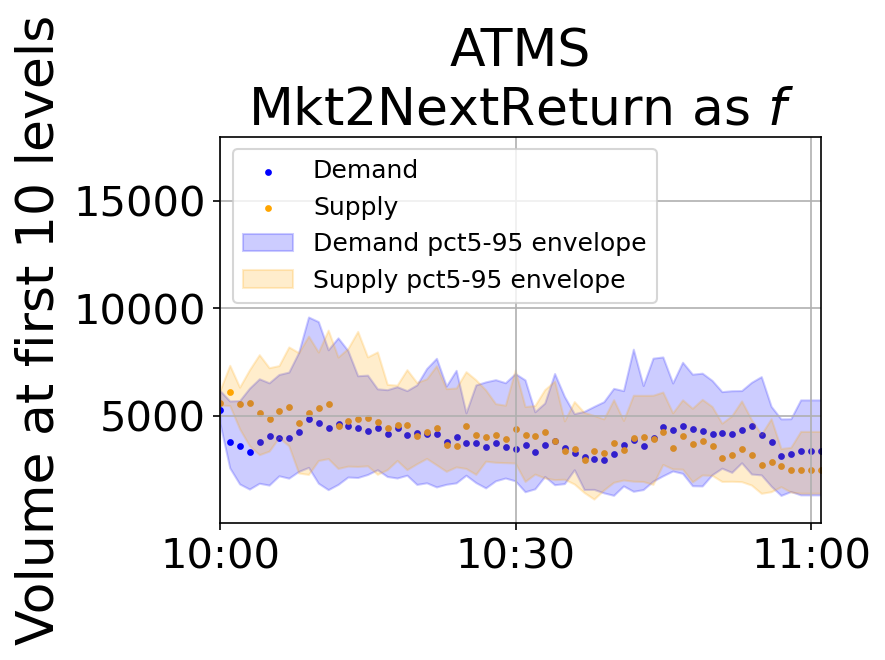}
\hspace{-0.06in}
\includegraphics[trim={1.3cm 0.25cm 0.25cm 0.25cm},clip,width = .195\textwidth]{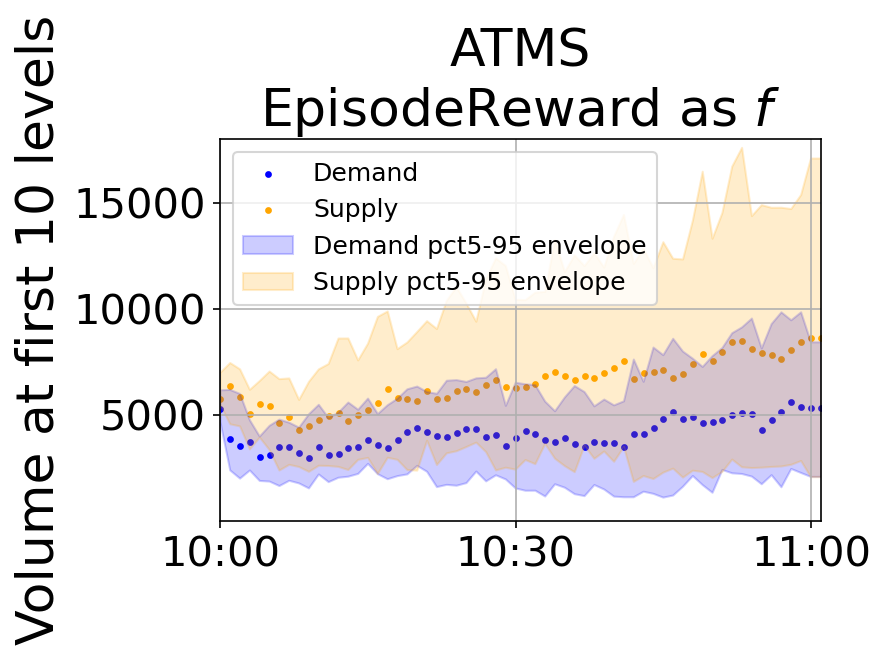}
\hspace{-0.06in}
\includegraphics[trim={1.3cm 0.25cm 0.25cm 0.25cm},clip,width = .195\textwidth]{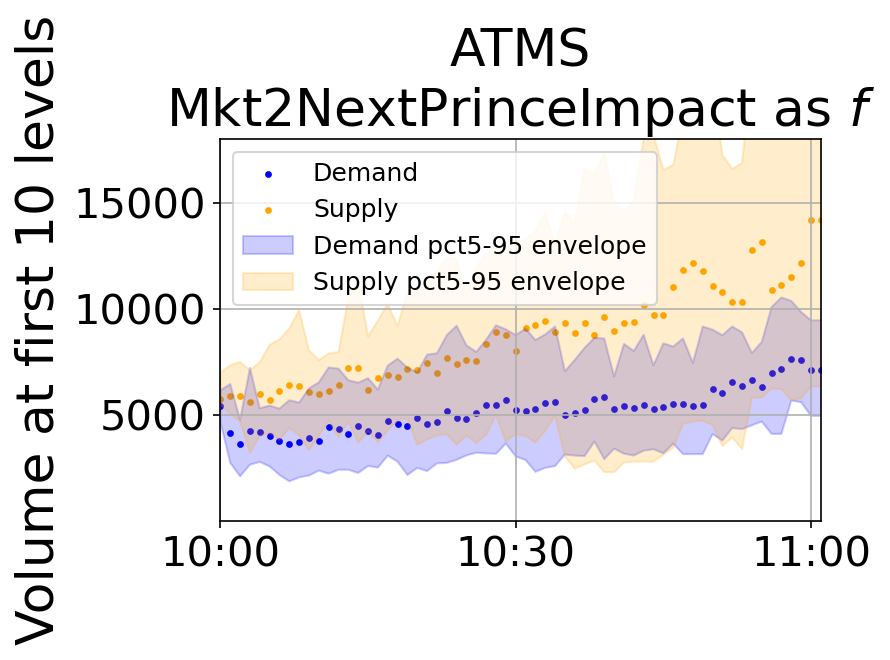}
\hspace{-0.06in}
\includegraphics[trim={1.3cm 0.25cm 0.25cm 0.25cm},clip,width = .195\textwidth]{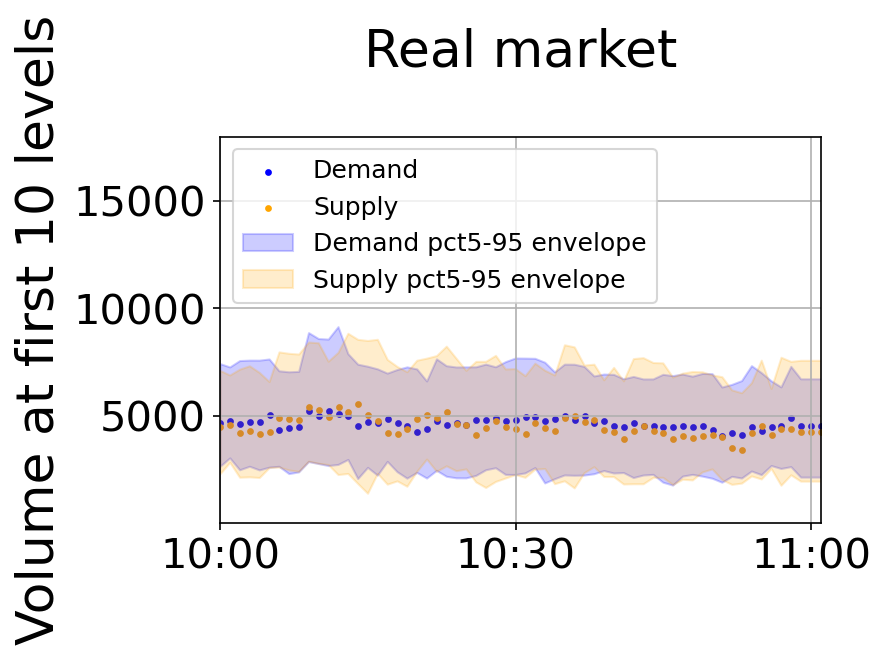}
}
\centerline{
\includegraphics[trim={.25cm 0.25cm 0.25cm 0.25cm},clip,width = .222\textwidth]{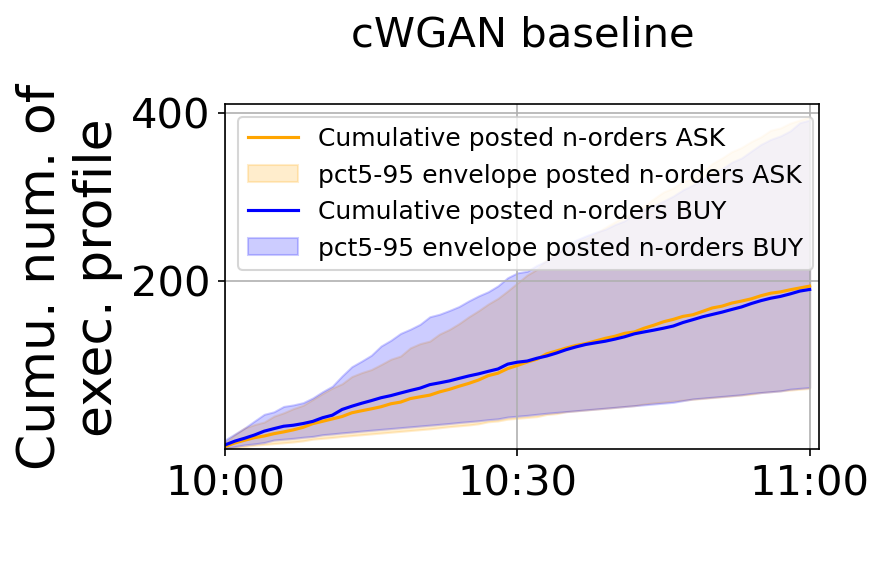}
\includegraphics[trim={2.25cm 0.25cm 0.25cm 0.25cm},clip,width = .19\textwidth]{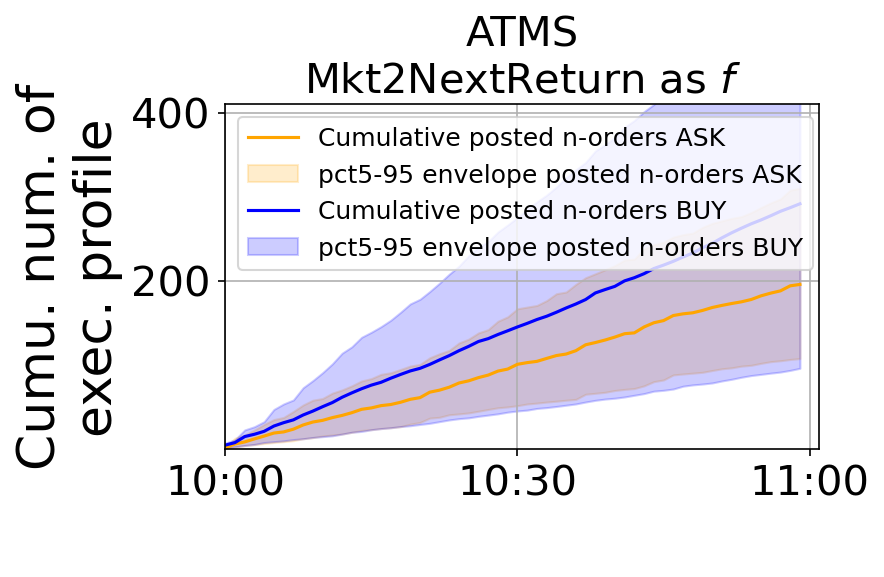}
\includegraphics[trim={2.25cm 0.25cm 0.25cm 0.25cm},clip,width = .19\textwidth]{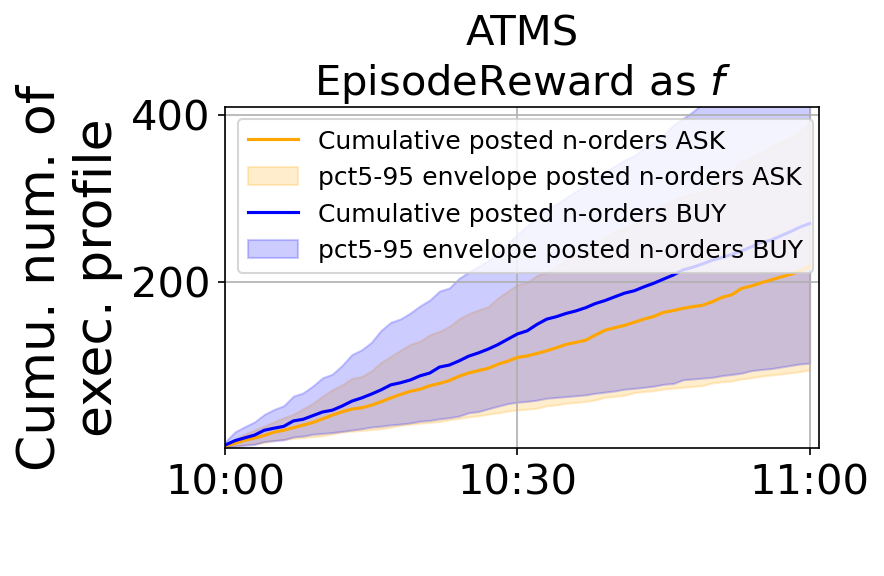}
\includegraphics[trim={2.25cm 0.25cm 0.25cm 0.25cm},clip,width = .19\textwidth]{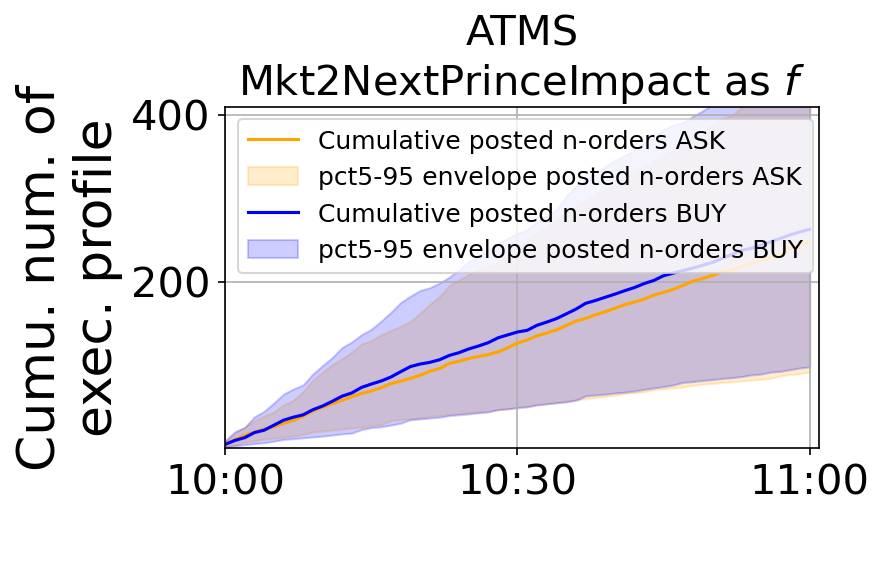}
\includegraphics[trim={2.25cm 0.25cm 0.25cm 0.25cm},clip,width = .19\textwidth]{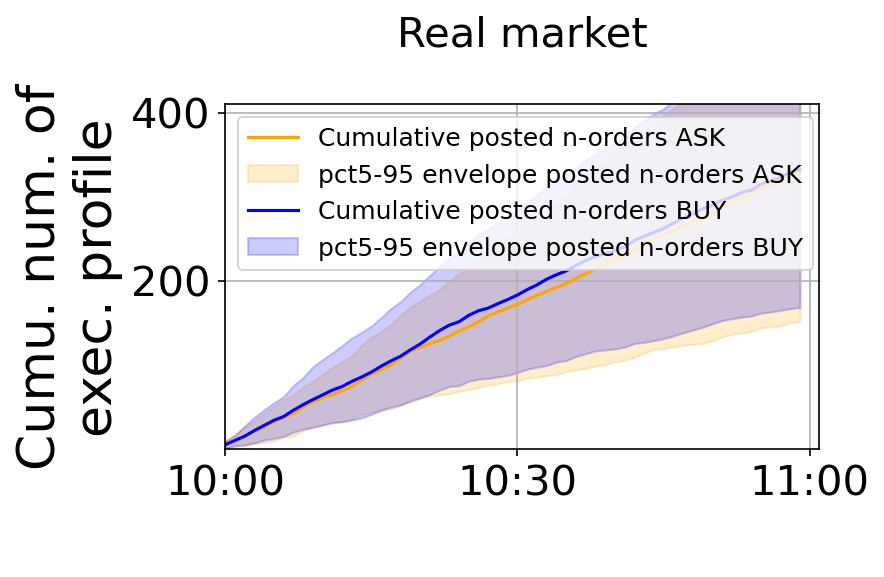}
}
\caption{{Comparison of stylized fact time series between our ATMS and cWGAN baseline. The top panel in the second column is exactly the third panel in Figure~\ref{fig:exp_effectiveness_1}, where ATMS optimizes the resemblance of \texttt{Mkt2NextReward} to reality and generates the most realistic \SFVten \  in terms of supply-demand match and the magnitude. Meanwhile, optimizing the resemblance of \texttt{Mkt2NextPriceImpact} (the fourth column) leads to improved match of \SFExecPrf \ to reality.}}
\label{fig:exp_comparison_1}
\end{figure*}

Specifically, we report the results of our proposed ATMS with different feedback choices in Figure~\ref{fig:exp_comparison_1}.
It is interesting to observe that using different $f$'s leads to improved similarity to reality for different stylized facts: From the first row of Figure~\ref{fig:exp_comparison_1}, we can see ATMS with \texttt{Mkt2NextReturn} as $f$ (second column in the figure) yields \SFVten \ most similar to reality in the sense that the supply and demand match with each other just like the real market, and their magnitudes are the most more similar to reality; Indeed, using \texttt{Mkt2NextReturn} as $f$ when training ATMS can lead to the best match of \SFVn \ ($n \in \{1, 5\}$) to reality as evidenced in Figure~\ref{fig:exp_comparison_more_volume} in Appendix~\ref{appendix:otherSF}.
Although ATMS with \texttt{Mkt2NextPriceImpact} as $f$ yields very different \SFVten \ from reality, it does achieve improved similarity compared to the initialization (see the first panel in Figure~\ref{fig:exp_effectiveness_1}) in terms of matching supply-demand. 
From the second row of Figure~\ref{fig:exp_comparison_1}, we can observe that ATMS with \texttt{Mkt2NextPriceImpact} as $f$ (fourth column in the figure) can yield the best \SFExecPrf: The improvement compared to \texttt{Mkt2NextReturn} is evident, whose BUY is significantly larger than SELL; Compared to cWGAN baseline, it correctly captures not only the magnitudes of SELL and BUY but also the pattern that BUY is slightly larger than SELL.
In addition, the third column in Figure~\ref{fig:exp_comparison_1} serves as ``direct evidence'' that \texttt{EpisodeReward} is not appropriate for training ATMS, reaffirming our previous claims.

\begin{figure}[!htp]
\centerline{
\includegraphics[trim={0.25cm 0.25cm 0.25cm 0.25cm},clip,width = .2168\textwidth]{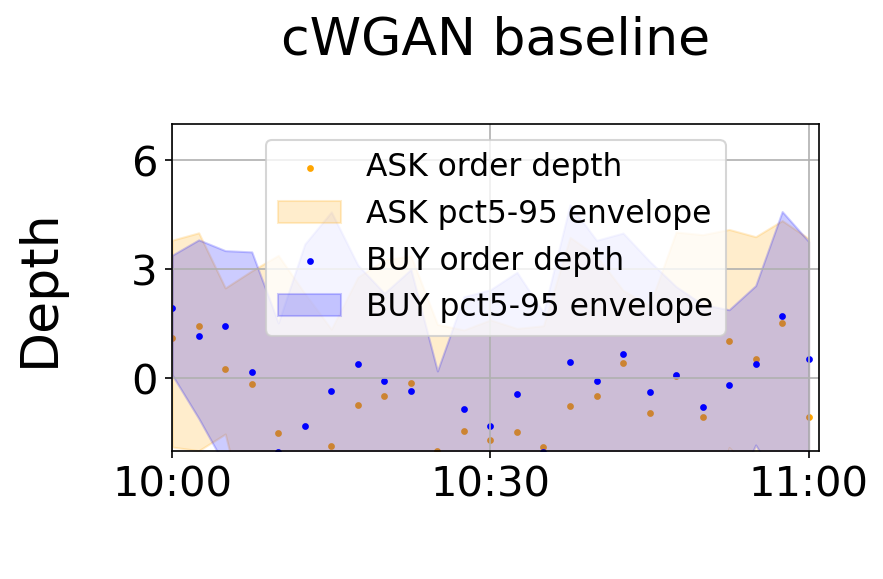}
\includegraphics[trim={2cm 0.25cm 0.25cm 0.25cm},clip,width = .19\textwidth]{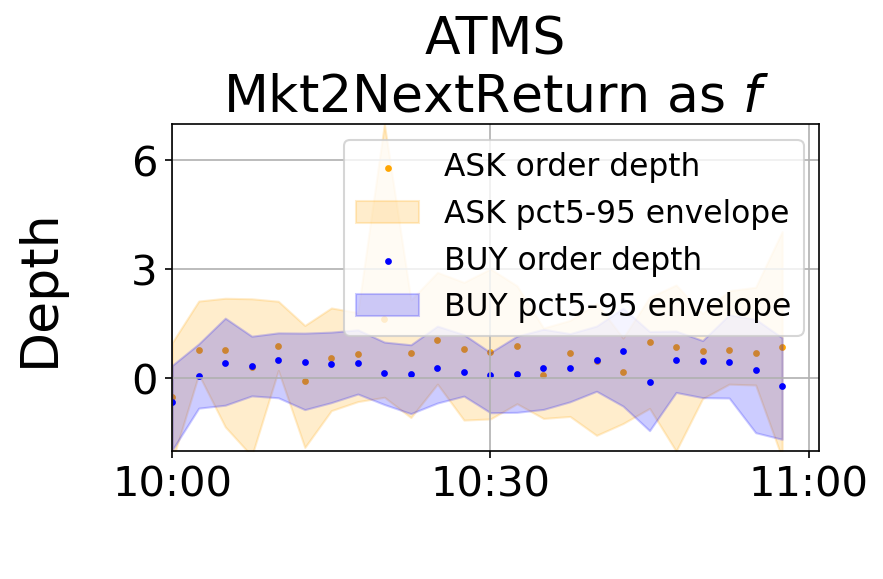}
\includegraphics[trim={2cm 0.25cm 0.25cm 0.25cm},clip,width = .19\textwidth]{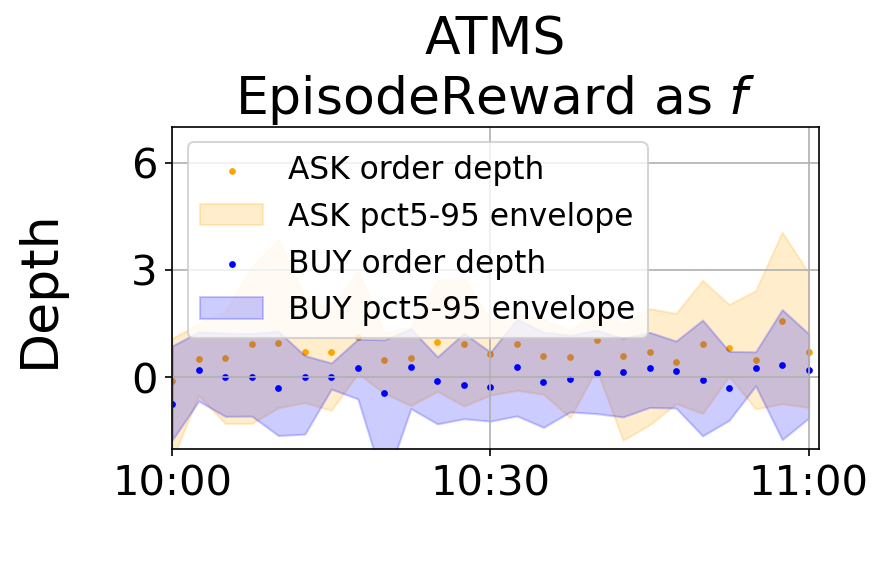}
\includegraphics[trim={2cm 0.25cm 0.25cm 0.25cm},clip,width = .19\textwidth]{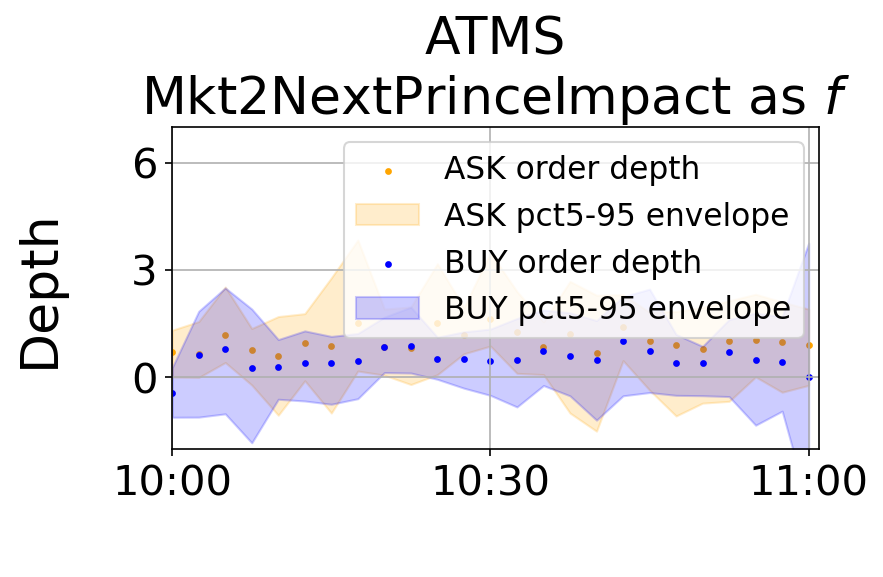}
\includegraphics[trim={2cm 0.25cm 0.25cm 0.25cm},clip,width = .19\textwidth]{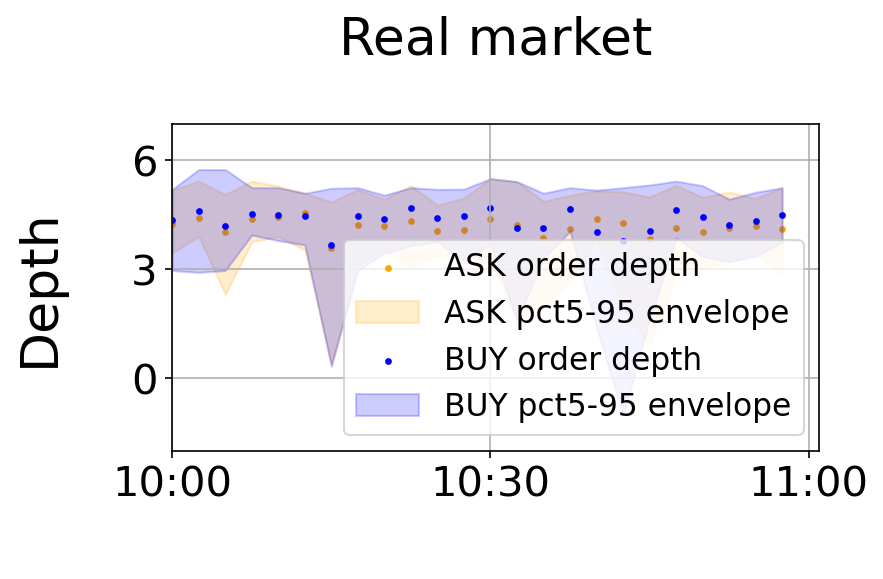}
}
\centerline{
\includegraphics[trim={0.25cm 0.25cm 0.25cm 0.25cm},clip,width = .2065\textwidth]{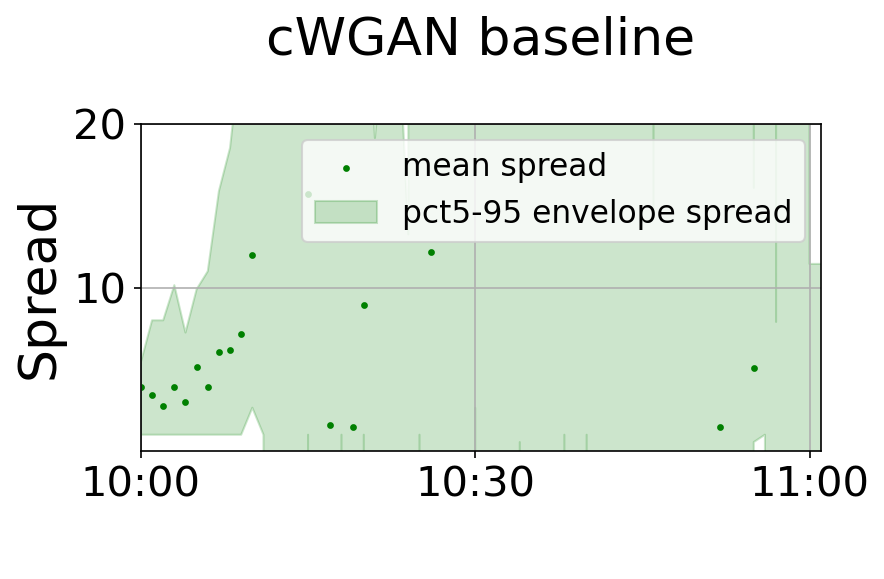}
\includegraphics[trim={1.3cm 0.25cm 0.25cm 0.25cm},clip,width = .19\textwidth]{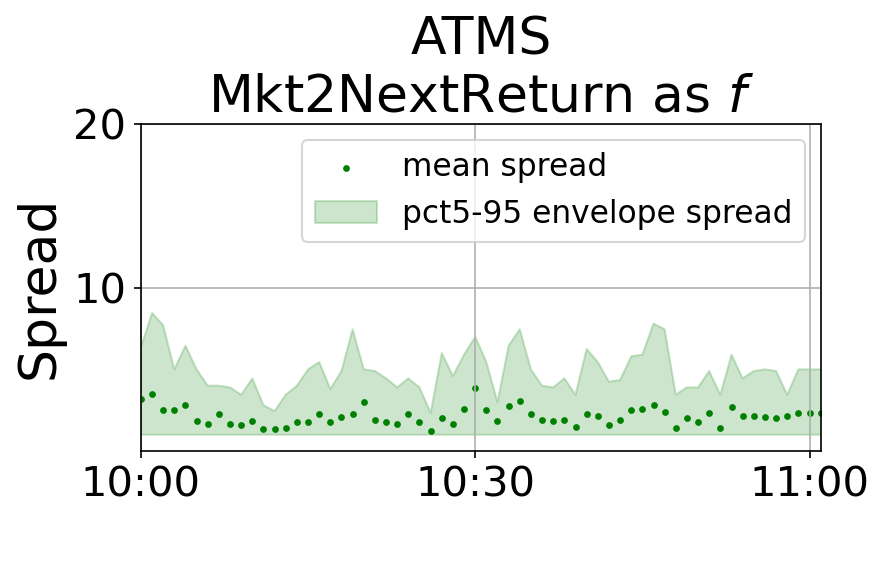}
\includegraphics[trim={1.3cm 0.25cm 0.25cm 0.25cm},clip,width = .19\textwidth]{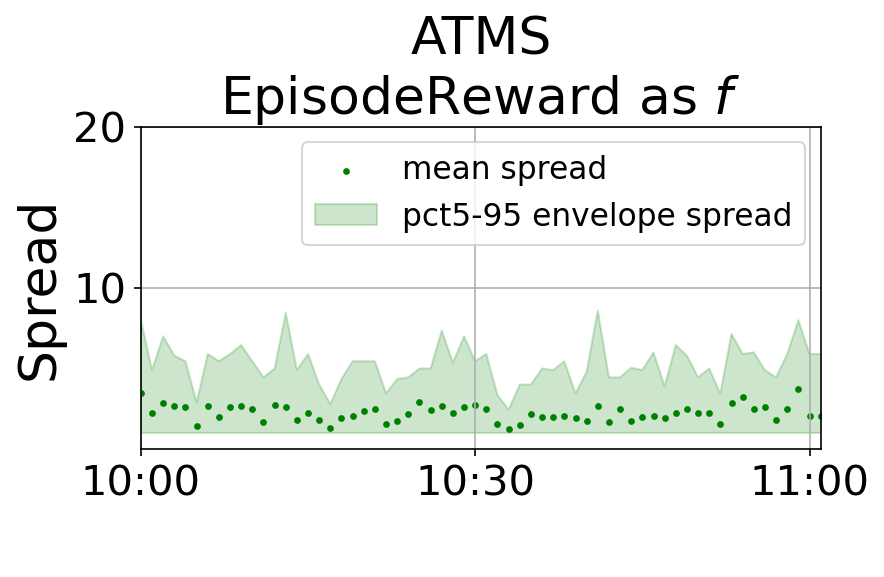}
\includegraphics[trim={1.3cm 0.25cm 0.25cm 0.25cm},clip,width = .19\textwidth]{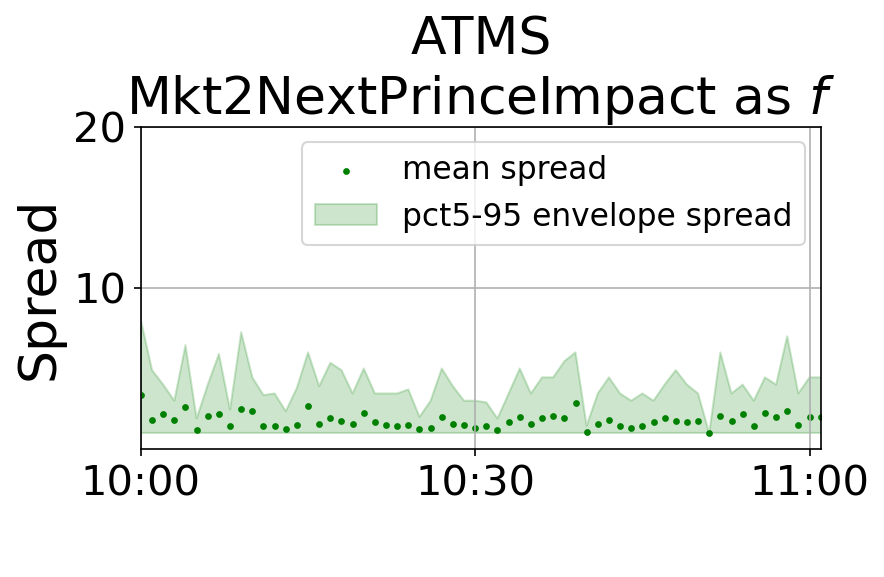}
\includegraphics[trim={1.3cm 0.25cm 0.25cm 0.25cm},clip,width = .19\textwidth]{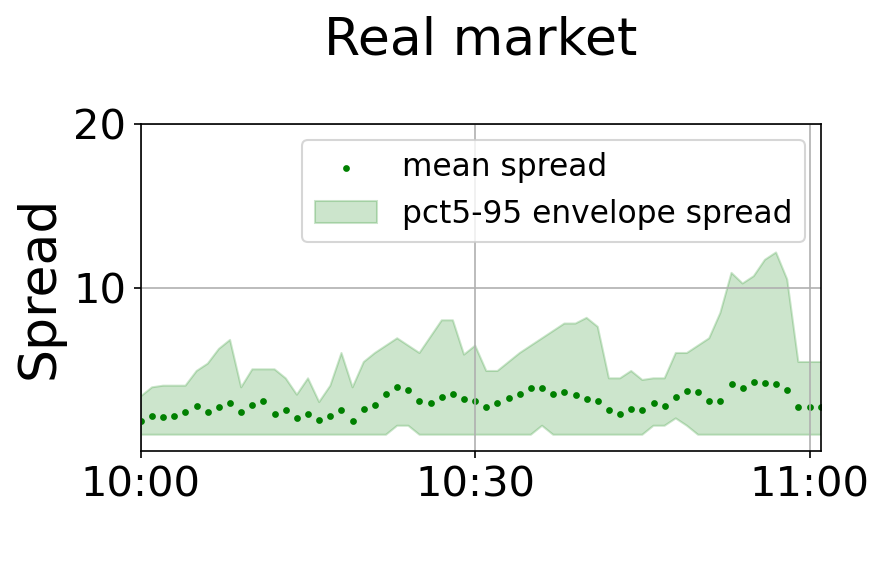}
}
\caption{Comparison of additional stylized facts (top: depth; bottom: spread) among different simulated markets (specified on top of each panel). We can observe that our ATMS with various feedback choices can better capture the variance of those stylized facts in real market compared to the cWGAN baseline.}
\label{fig:exp_comparison_others_same}
\vspace{0.3in}
\centerline{
\includegraphics[trim={0.25cm 0.25cm 0.25cm 0.25cm},clip,width = .2065\textwidth]{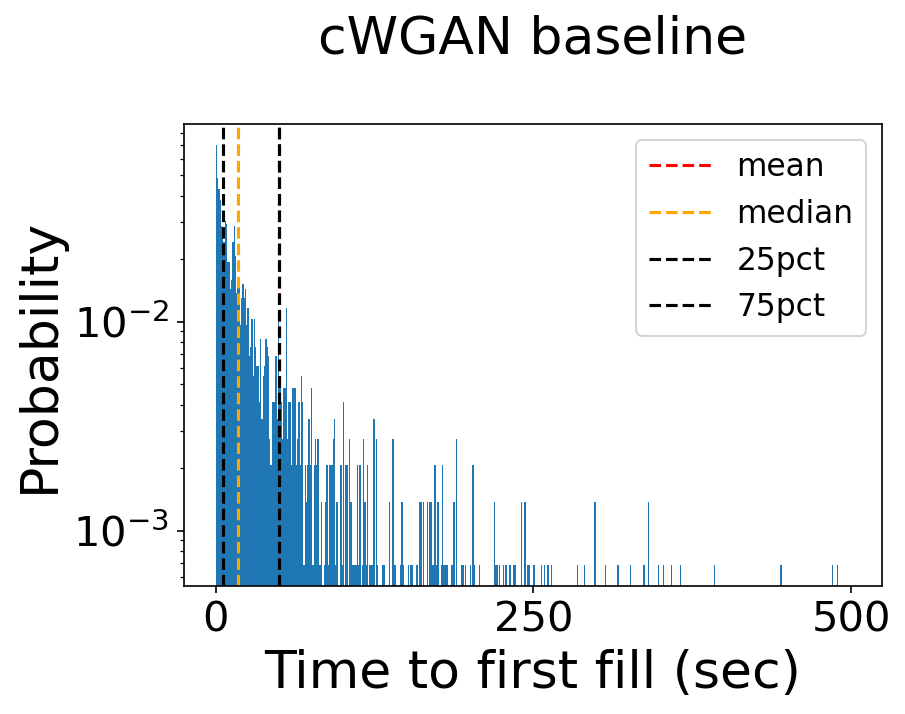}
\includegraphics[trim={1.3cm 0.25cm 0.25cm 0.25cm},clip,width = .19\textwidth]{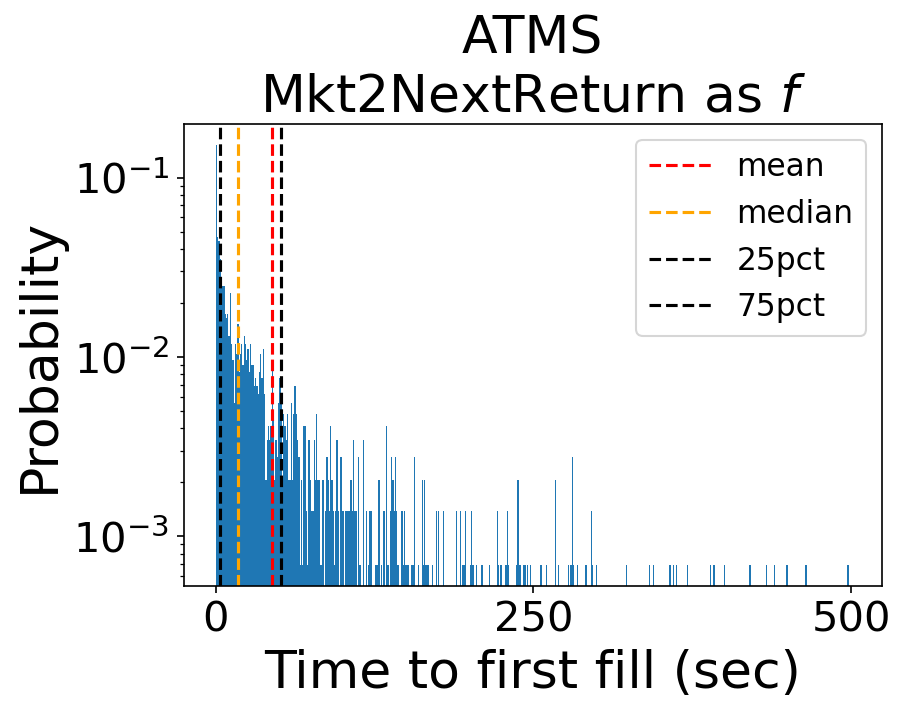}
\includegraphics[trim={1.3cm 0.25cm 0.25cm 0.25cm},clip,width = .19\textwidth]{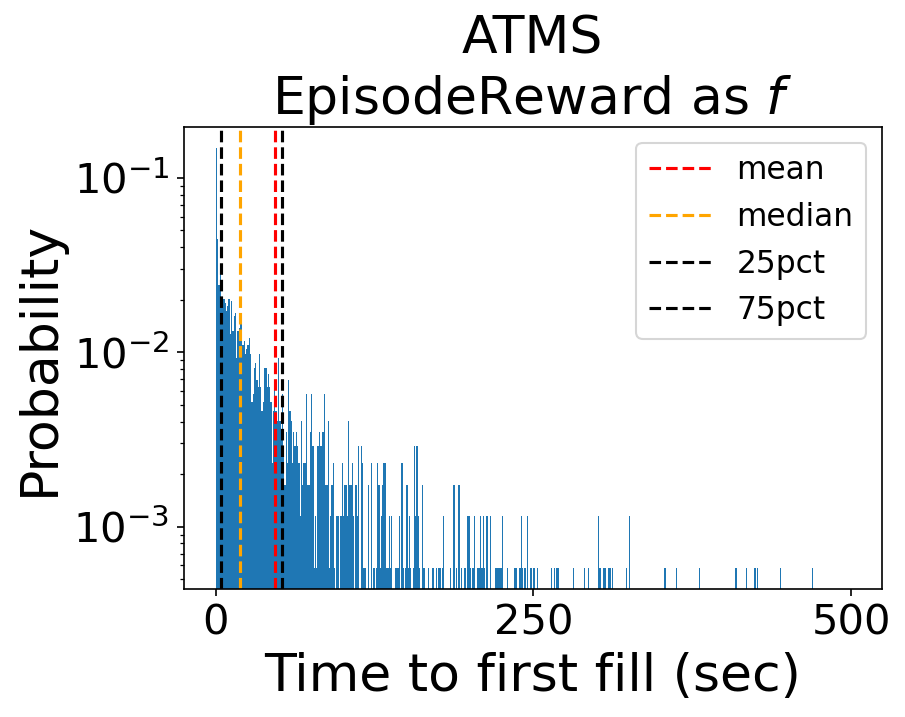}
\includegraphics[trim={1.3cm 0.25cm 0.25cm 0.25cm},clip,width = .19\textwidth]{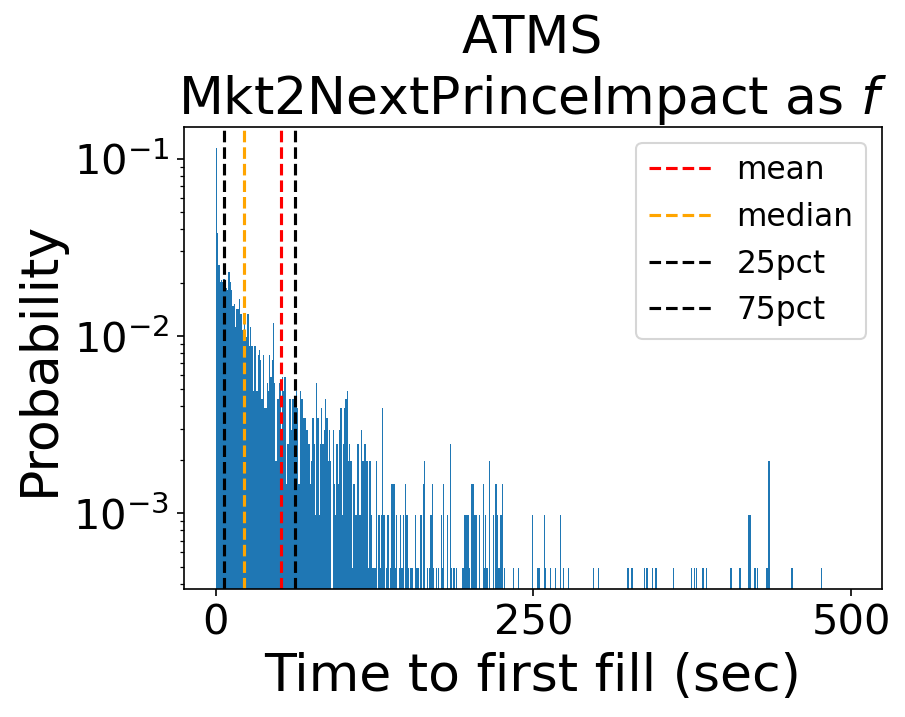}
\includegraphics[trim={1.3cm 0.25cm 0.25cm 0.25cm},clip,width = .19\textwidth]{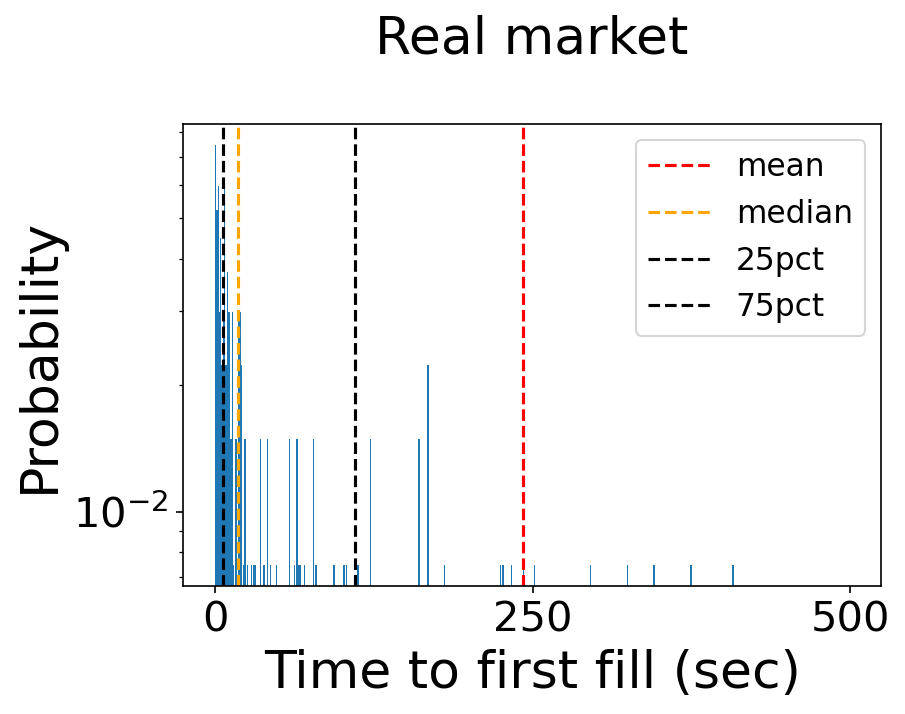}
}
\vspace{0.08in}
\centerline{
\includegraphics[trim={0.25cm 0.25cm 0.25cm 0.25cm},clip,width = .222\textwidth]{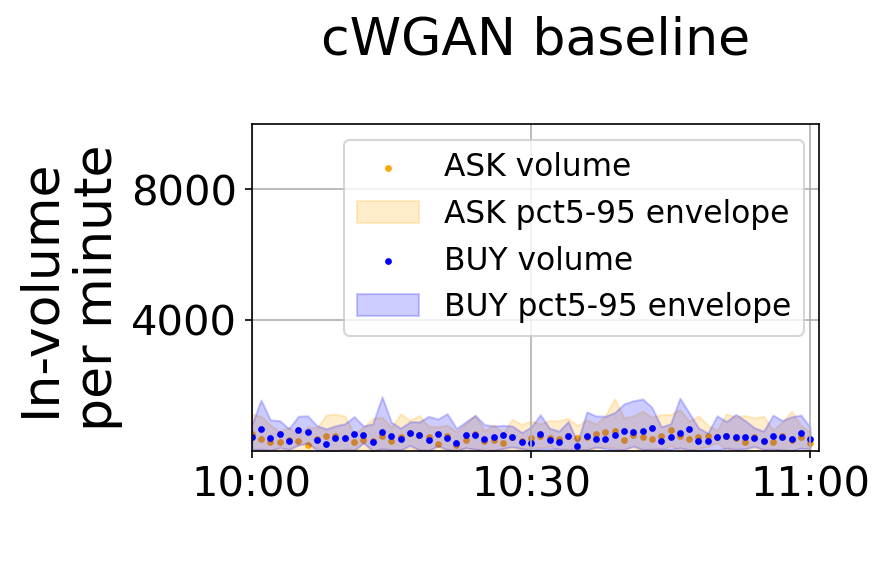}
\includegraphics[trim={2.25cm 0.25cm 0.25cm 0.25cm},clip,width = .19\textwidth]{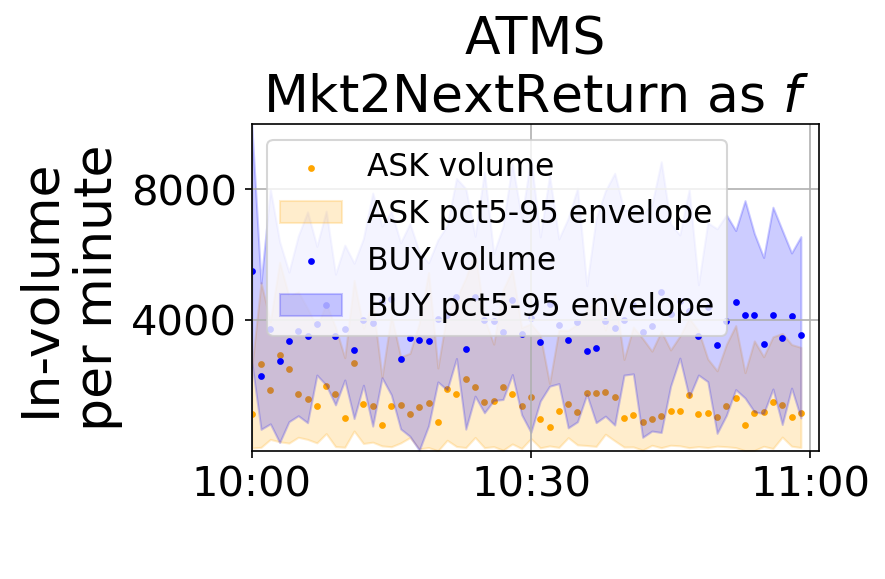}
\includegraphics[trim={2.25cm 0.25cm 0.25cm 0.25cm},clip,width = .19\textwidth]{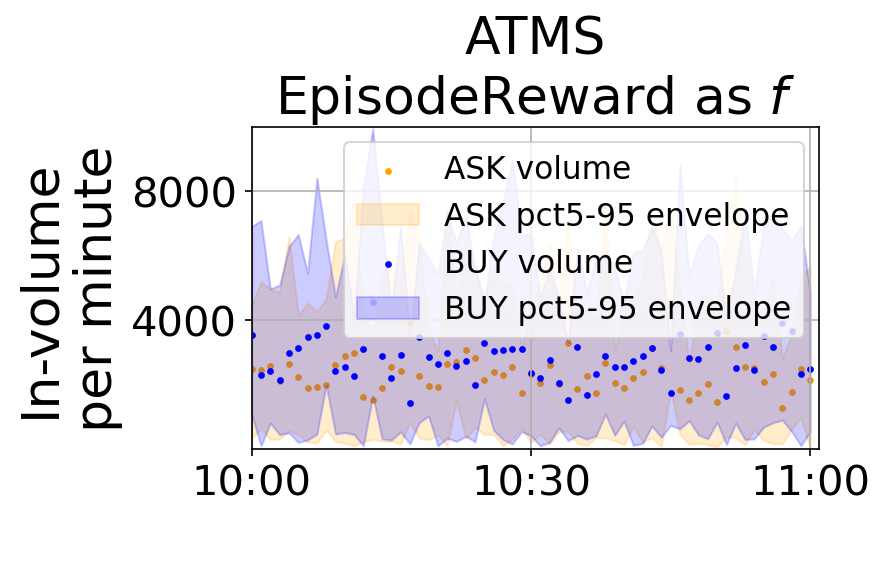}
\includegraphics[trim={2.25cm 0.25cm 0.25cm 0.25cm},clip,width = .19\textwidth]{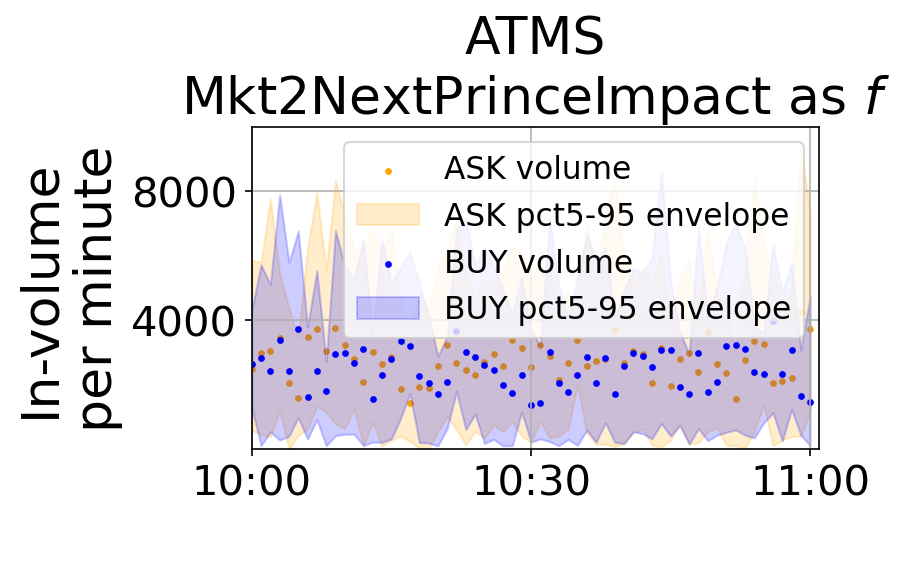}
\includegraphics[trim={2.25cm 0.25cm 0.25cm 0.25cm},clip,width = .19\textwidth]{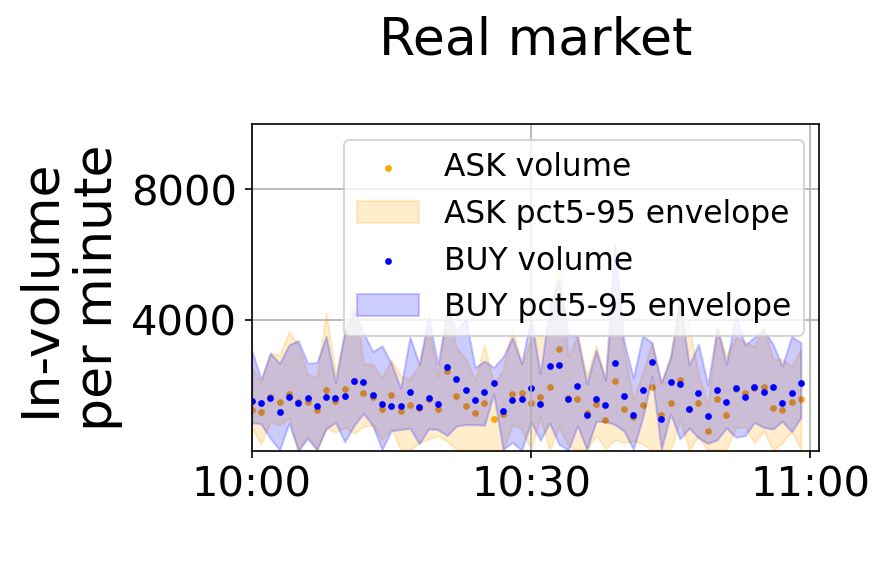}
}
\caption{Comparison of other stylized facts (top: time to first fill; bottom: in-volume per minute) among different simulated markets (specified on top of each panel). We can observe that neither our ATMS nor the cWGAN baseline can correctly capture those stylized facts in real market, suggesting future work to further improve the market simulator.}
\label{fig:exp_comparison_others_diff}
\end{figure}

For completeness, we also report results for other stylized facts (e.g., spread) in Figures~\ref{fig:exp_comparison_others_same} and~\ref{fig:exp_comparison_others_diff}, from which we can observe that there is still a gap between simulated and real markets, and neither our ATMS nor cWGAN can generate market data with those stylized facts similar to reality. Nevertheless, we can observe from Figure~\ref{fig:exp_comparison_others_same} that our ATMS have better variance matching to reality compared to the cWGAN baseline for depth and spread, even though all of them do not correctly capture the magnitude. All of the above results show that our ATMS with properly chosen feedback (or rather, the application of our INTAGS to trading markets) can generate much more realistic data compared to the cWGAN baseline.

\section{Conclusion and Discussion}\label{sec:discussion}

This work proposes a MAS distance metric, on which we build INTAGS online training framework. The effectiveness is demonstrated by the market simulation application, reaffirming our claim that incorporating the agents' live interactions is of vital importance for simulating environments with interactive agents. Our experiments demonstrate the potential of INTAGS in real production; As illustrated in Figure~\ref{fig:all_illus}, INTAGS only needs a fixed collection of real feedback $f(\boldsymbol{\cT}_{\pi, \real})$ \eqref{eq:realrollout}, i.e., historical reply. Thus, the deployment of our INTAGS, such as ATMS in market simulation, does not interfere the real production. 
However, since INTAGS's deployment requires the Exp agent to take real-world strategy, which typically targets very complex tasks on a large time horizon $T$, one potential limitation is the computational complexity: Each update of parameter $\theta$ in Algorithm~\ref{alg:algorithm} requires $\cO(NbT)$ interactions, resulting in $\cO(NbT^2)$ steps. Luckily, the complexity can be reduced to $\cO( N b T_0 T)$ by approximating the gradient as:


\begin{equation*}
\nabla_\theta \cL \simeq \sum_{t=1}^{\boldsymbol{T_0}} \EE_{x \sim p_\theta\left(\cdot |  S_t^{\tau_t}\right)}\left[\nabla_\theta \log p_\theta\left(A_t |  S_t^{\tau_t}\right) Q_f\left(\Tilde{\cT}_{\pi, \theta}^{1:t}, A_t\right)\right].
\end{equation*}


\noindent
This is because $T_0$ (and $b$) determines how many BG agent actions are sampled and penalized, and thus can be treated as batch-size in stochastic GD; This makes $T_0$ a tunable hyperparameter; By carefully tuning $T_0$ (and $b$; see the hyperparameter selection in Appendix~\ref{appendix:training}), we can finish the INTAGS training within a reasonable time frame (around $2$ hours) using Amazon Web Services (r6i.24xlarge, 96 CPUs, 768 GiB memory). In practice, after proper selection of $N$\footnote{$N$ controls how well the estimated metric $Q_f$ is, and it has already been tuned by experiments in Section~\ref{sec:loss_exp} --- We want to find the smallest $N$ to meet Ax2 and guarantee an effective metric (i.e., the estimated $Q_f$ can differentiate simulated and real Envs).}, $b, T_0$, the complexity is merely linear w.r.t. the horizon $T$, which is determined by the real tasks and thus untunable.

\section*{Acknowledgment}
The authors would like to thank Haibei Zhu for his valuable suggestion on algorithmic trading strategy construction, and Yousef El-Laham and Penghang Liu for their advice on the method presentation.

\section*{Disclaimer}
This paper was prepared for informational purposes by
the Artificial Intelligence Research group of JPMorgan Chase \& Co. and its affiliates (``JP Morgan''),
and is not a product of the Research Department of JP Morgan.
JP Morgan makes no representation and warranty whatsoever and disclaims all liability,
for the completeness, accuracy or reliability of the information contained herein.
This document is not intended as investment research or investment advice, or a recommendation,
offer or solicitation for the purchase or sale of any security, financial instrument, financial product or service,
or to be used in any way for evaluating the merits of participating in any transaction,
and shall not constitute a solicitation under any jurisdiction or to any person,
if such solicitation under such jurisdiction or to such person would be unlawful.

\bibliographystyle{plainnat}  
\bibliography{ref}

\appendices

\addcontentsline{toc}{section}{Appendix} 
\part{\centering \Large Appendix of \\ \papertitle} 

\topskip0pt

\parttoc 


\newpage

\section{Extended Literature Survey}\label{appendix:literature}
Here, we extend our literature survey to more relevant topics.

\paragraph{Generative models in finance.}
Training experimental agents usually involves interaction with the real market, which is often impractical. To create a suitable market for the experimental agent, generative models representing all background agents are needed and have gained prominence in finance, with GAN being a particularly popular choice. 
To the knowledge of the author, this line of research traces back to \citet{kumar2018ecommercegan,shi2019virtual}, who adapted GAN to generate BUY orders in e-commerce markets. Notably, GAN has been utilized to model more complex stock markets and simulate various types of stock market data, including transaction event time \citep{xiao2017wasserstein,xiao2018learning}, price \citep{zhang2019stock,wiese2020quant,koshiyama2021generative}, and even orders \citep{li2020generating,coletta2021towards,coletta2022learning}.

\paragraph{Generative models for RL.}
Our proposed INTAGS is closely related to RL: On one hand, the generator is formulated as a stochastic RL policy and trained via policy gradient. On the other hand, the downstream task --- the Exp agent --- can be done by RL. In literature, reinforcement learning and generative models have been traditionally considered separate fields until some recent developments reveal their promising connections. One notable application involves leveraging generative models in RL training, particularly relevant in the context of optimal execution tasks in financial markets. Indeed, it is a popular approach to study financial tasks with an interactive simulator: For example, \citet{karpe2020multi} trained RL agent to perform optimal execution tasks by interacting with parametric market simulator ABIDES \citep{byrd2020abides,amrouni2021abides} such that the training process captures real-world dynamics, enabling agents to grasp the consequences of their actions on the responses of other market participants; 
\citet{kuo2021improving,koshiyama2021generative} trained agents to perform downstream financial tasks with the help of a conditional GAN-based market simulator.
One notable work studying the connection between GAN and RL is \citet{ho2016generative}, which proposed to mimic the expert policy from instances by inverse RL (IRL) followed by RL; They theoretically proved that explicitly learning of the cost function in the IRL could be bypassed, which enabled end-to-end learning of the policy from the expert policy instances. Moreover, the resulting imitation learning problem takes a GAN formulation, bridging RL and GAN from a very novel perspective. 
Other contributions in this direction include 
\citet{finn2016connection, fu2017learning}, who formulated IRL as GAN problem.

\paragraph{RL for Generative models.}
As pointed out by \citet{yu2017seqgan,shi2019virtual} as well as our work, the generator can be formulated as the (RL) agent policy, and therefore imitation learning can be used for market simulation, which opens up more possibilities in this area (i.e., market simulation).
However, it is difficult to train adversarial imitation learning models in practice since careful reward augmentation \citep{bhattacharyya2019simulating} and curriculum design \citep{behbahani2019learning} are needed in practice, and this might explain why GAN is still the most popular approach for market simulation. We refer readers to \citet{torabi2019recent} for a recent survey on imitation learning.
On the other hand, as most GAN-based simulators, \citet{ho2016generative} also used the classic loss function and this is the key difference from our work: the proposed interactive agent-based environment distance metric and the derived training framework --- INTAGS --- are the main contribution of our work. 
Another seemingly closely related work in this direction is
\citet{ruizlearning}, who studied the classification problem in the presence of limited real data via a classifier trained on generative models.
They claimed to leverage RL to help with the GAN training.
However, as pointed out by the reviewer, even though the feedback from evaluating the classifier on the real data can help improve the generative model, there is no clear state and action space definition nor sequential decision-making, rendering the claim of reformulating the GAN training as RL less convincing.
Other works that leveraged RL to help with the GAN training include:
\citet{sarmad2019rl}, who proposed to use RL to control the input to generators in the GAN, and
\citet{tian2020off}, who proposed to use RL to help search for the optimal GAN architecture.
However, non of those works leverage sequential agents' interactions to develop a metric for the training of a generative model, which is the key difference between our work and those existing works.

\section{Additional Details of INTAGS}\label{appendeix:method}

\subsection{Causal effect estimation for feedback}\label{appendix:causal}

One of our main contributions is the novel interactive agent-based MAS metric, which compares the discrepancy of the feedback (empirical) distributions under different environments. The feedback is defined over a longer sequential state-action-next state chain, compared to the previous state-action pair used in classic cWGAN. By considering a longer chain, our metric considers and introduces a much more complex sequential dependency.

To understand what is the complex sequential dependency, let us first recall one proposed feedback in the market simulation application: the \texttt{Mkt2NextReturn}; That is, the feedback is the causal effect from Exp AT agent placing market (BUY) order to the next market return. However, at time step $t$, the market evolution $s_t$ is not only the result of Exp AT agent's action $a_t$ and the BG agents' responses $A_t^j$'s, but also {\it correlated} to previous market return $s_{t-1}$ --- this is the additional sequential dependency introduced by our feedback; in the classic setting, the BG agents' state-action pairs $(S_t^j,A_t^j)$'s have very clear and simple dependency that action $A_t^j$ only depends on the input market state $S_t^j$.

Next, let us elaborate on why introducing such additional sequential dependency poses a challenge in feedback estimation. Again let us consider feedback \texttt{Mkt2NextReturn} in the market simulation; To estimate $f$ from collected sequential observations, one can {\it select the time steps where $a_t = {\rm placing \ market \ order }$} and take the average of corresponding $s_t$'s as the estimator. To be precise, the naive estimator of feedback $f$, i.e., the causal effect from Exp agent action to the next state, say $j$-th element of the state vector, is defined as
\[{f}_{\rm naive} =  \frac{1}{\#\{t: a_t = {\rm placing \ market \ order }\}} \sum_{t = 1}^T {s_{t}(j) \mathbf{1}_{\{a_t = {\rm placing \ market \ order }\}}},\]
where $\#$ denotes the cardinality of a set, $s_{t}(j)$ is the $j$-th element of the state vector (e.g., return), and $\mathbf{1}$ is the indicator function. 

The problem of the above naive estimator is the {\it selection bias}, i.e., the selected sub-population is not representative enough of the whole population, making the average effect estimator above a biased one; To understand this, let us consider a very simple example, where we want to study the effect of carrying a lighter to developing lung cancer: If the effect from smoking, which acts as a common cause to both carrying a lighter and developing lung cancer, is not considered, a false causal conclusion (which is indeed just correlation) that carrying a lighter will result in lung cancer will be made. Under the potential outcome framework by \citet{rubin1974estimating}, the causal effect estimation from the treatment variable to the outcome variable needs to take into account the effect of pre-treatment covariates, called potential confounders. 
There are two solutions to this problem: 
\begin{itemize}
    \item \textbf{(Apch1)} One naive approach is to manually break the correlation between $s_{t-1}$ and $a_t$; In the market simulation, we can use zero-intelligent Exp AT agent that aggressively or randomly place orders; However, as we mentioned in our discussion (see Section~\ref{sec:discussion}), the deployment of INTAGS needs the Exp agent to be the real one in the production system and therefore it is unlikely that such an Exp agent is zero-intelligent.
    \item \textbf{(Apch2)} One very popular approach to estimate causal effect from observational data is to leverage inverse probability weighted (IPW) estimator \citep{horvitz1952generalization} to adjust for the potential confounders; The high-level idea is to adjust the weight of each selected sample such that the re-weighted sub-population is the same with the whole population in theory, addressing the issue of selection bias. To be precise, we denote the propensity score, which is the probability of ``receiving treatment'', as
\[ e(s) = \PP( a = {\rm placing \ market \ order } \mid {\rm state } = s) = \pi(s, a),\]
where $\pi$ is the policy of the Exp agent (or AT agent in our market simulation application).
The IPW estimator of feedback $f$ is defined as:
\[{f}_{\rm IPW} = \frac{1}{\#\{t: a_t = {\rm placing \ market \ order }\}} \sum_{t = 1}^T \frac{s_{t}(j) \mathbf{1}_{\{a_t = {\rm placing \ market \ order }\}}}{\hat e(s_{t-1})},\]
where $\hat e$ denotes the estimated propensity score (typically through logistic regression).
In practice, to ensure estimates $\hat e$'s are not overly small, they are typically clipped using a certain threshold (denoted by
${\rm PSthres}$), i.e., $\hat e \leftarrow \max \{\hat e, \rm PSthres\}$.

\end{itemize}



\subsection{Value function evaluation}\label{appendix:value_eval} 
The value function $Q_f$ estimation relies on $N'$ feedbacks from the real Env and $N$ feedbacks obtained by performing $N$ complete rollouts under the world Env.
\begin{figure}[!htp]
\centerline{
\includegraphics[width = .75\textwidth]{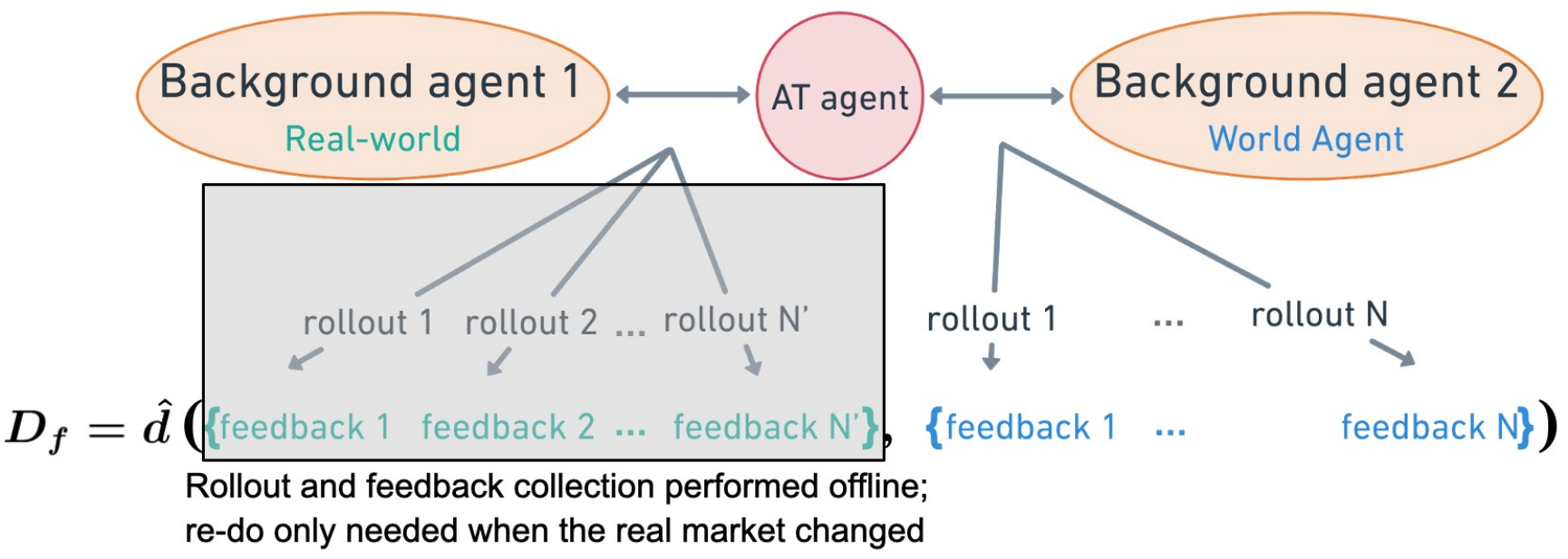}
}
\caption{Illustration of the calculation of our proposed interactive agent-based metric.}
\label{fig:metric_illus}
\end{figure}
As illustrated in Figure~\ref{fig:metric_illus}, the interaction under real Env to obtain the feedback only needs to be performed offline; During the training of the world BG agent policy/generator network in Algorithm~\ref{alg:algorithm}, the interaction under world Env and the network parameter update are performed iteratively.

In this work, the probability distribution distance metrics we consider are MMD, ED, and EMD. Other discrepancy metrics such as Kullback–Leibler divergence and Jensen–Shannon divergence are less favorable as we can observe the supports of two empirical probability distributions are different in Figures~\ref{fig:exp2-1}, \ref{fig:exp2-3} and \ref{fig:exp2-4} (especially when the number of rollouts is small). 

Given two sets of samples $\{f_1,\dots,f_N\}$ and $\{g_1,\dots,g_{N'}\}$, an unbiased estimator of MMD can be obtained through the following U-statistic \citep{gretton2012kernel}:
\begin{equation*}
    \begin{split}
        \hat d_{\rm MMD}(\{f_1,\dots,f_N\}, \{g_1,\dots,g_{N'}\}) =
        \frac{1}{N(N-1)} \sum_{i=1}^{N} \sum_{j\neq i}^{N} k(f_i, f_j) - \frac{2}{N N'} \sum_{i=1}^{N} \sum_{j=1}^{N'} k(f_i, g_j) \\
        + \frac{1}{N'(N'-1)} \sum_{i=1}^{N'} \sum_{j\neq i}^{N'} k(g_i, g_j),
    \end{split}
\end{equation*}
where $k(\cdot,\cdot)$ is the user-specified kernel function. Popular choices include Gaussian kernel with bandwidth parameter $\sigma > 0$, 
\[k(f_1,g_1) = \exp\{ \| f_1-g_1 \|_2^2 /(2\sigma^2)\},\]
where $\| f_1-g_1 \|_2$ represents the Euclidean distance.

In addition, the empirical estimate of Energy Distance is as follows:
\begin{equation*}
    \begin{split}
        \hat d_{\rm ED}(\{f_1,\dots,f_N\}, \{g_1,\dots,g_{N'}\}) = \frac{1}{N(N-1)} \sum_{i=1}^{N} \sum_{j\neq i}^{N} \| f_i - f_j \|_2^2 - \frac{2}{N N'} \sum_{i=1}^{N} \sum_{j=1}^{N'} \| f_i - g_j \|_2^2 \\ + \frac{1}{N'(N'-1)} \sum_{i=1}^{N'} \sum_{j\neq i}^{N'} \| g_i - g_j \|_2^2.
    \end{split}
\end{equation*}
It is worthwhile noting that ED is a special case of MMD with linear kernel; Indeed, their equivalence has already been established \citep{sejdinovic2013equivalence}. Lastly, the estimation of Earth Mover's Distance is done by a popular built-in implementation in \texttt{wasserstein\_distance} in \texttt{scipy.stats}; We omit further details as EMD is a less favorable $\hat d$ choice according to our empirical evidence.

\subsection{Density estimation}\label{appendix:KDE} 
Here, since the generator network $G_\theta$ is a deterministic transformation, we apply kernel density estimation (KDE) to obtain the induced density function $p_\theta$; We choose KDE since it enjoys a closed-form and differentiable solution. In our experiment, we use a Gaussian RBF kernel with a bandwidth parameter selected by the median heuristic (which is referred to as adaptive bandwidth). Note that using KDE in the objective function is not novel in literature, and there exists work that proposed to include the bandwidth as a learn-able parameter during the optimization \citep{gomez2020kernel}, which is referred to as optimized bandwidth. Alternatively, one can choose a fixed bandwidth parameter in KDE; Here, since the bandwidth choice is not a major contribution in our work, we leave the performance of those bandwidth choices (i.e., adaptive, fixed or optimized; this work takes the adaptive one) as future development and the user can freely choose among them during the implementation.

\subsection{Training algorithm: a graphical illustration}\label{appendix:gradient}
\begin{figure}[!htp]
\centerline{
\includegraphics[width = .75\textwidth]{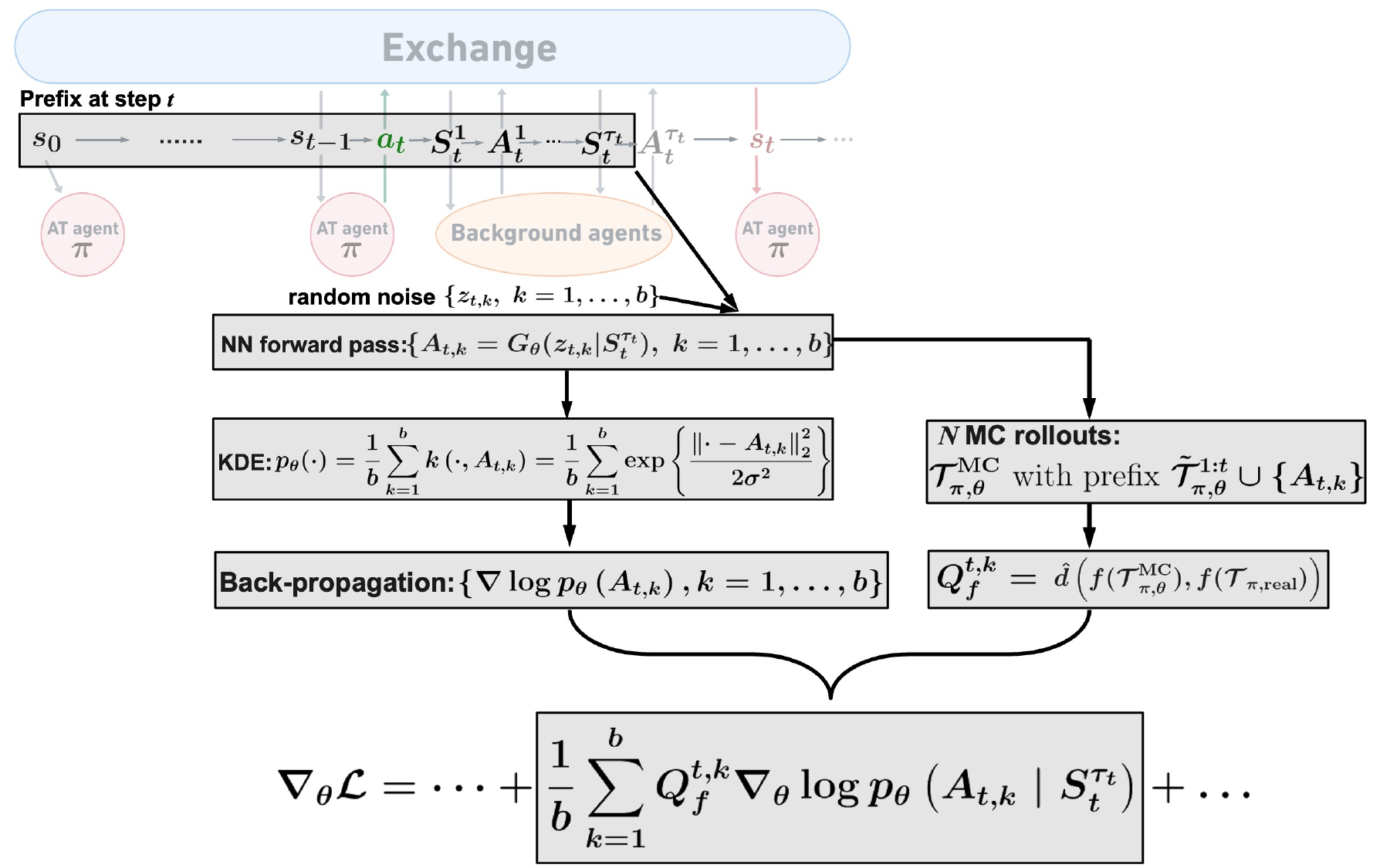}
}
\caption{Illustration of Algorithm~\ref{alg:algorithm} --- the forward-backward pass and and the gradient evaluation.}
\label{fig:forward_backward_illus}
\end{figure}

In this part, we use graphical illustration to show how to perform Algorithm~\ref{alg:algorithm} in practice. In particular, we illustrate the forward pass and back-propagation to obtain the gradient in Figure~\ref{fig:forward_backward_illus}.

\section{Additional Background Knowledge of Market Simulation Experiment}

\subsection{Limit order book}\label{appendix:LOB}
\begin{figure}[!htp]
\centerline{
\includegraphics[width = \textwidth]{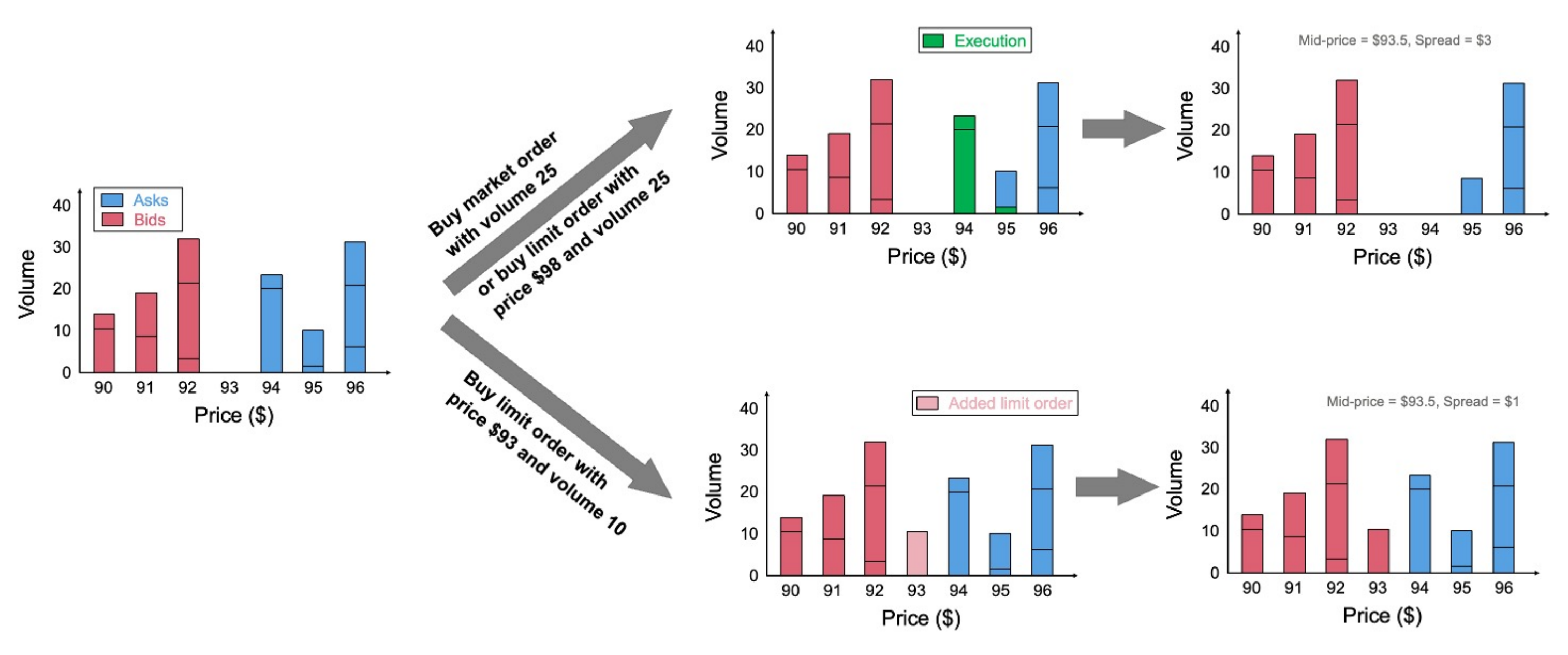}
}
\caption{Illustration of the dynamics of LOB.}
\label{fig:LOB_dynamics_illus}
\end{figure}

\paragraph{LOB dynamics.}
We begin with introducing the dynamics of LOB, as shown in Figure~\ref{fig:LOB_dynamics_illus};
We illustrate this dynamics of the LOB with three cases:
\begin{itemize}
    \item Case 1: A buy market order with volume 25 arrives, leading to 25 shares of order executed on ask side based on price and order arrive time. The execution price will be the ask price, and the resulting LOB market will have mid-price and spread shifted to $\$ 93.5$ and $\$ 3$.
    \item Case 2: A buy limit order with price $\$ 98$ and volume 25 arrives. As $\$ 98$ exceeds the current ask price (i.e., $\$ 94$), the incoming will be executed at the ask price. The resulting LOB market will be exactly the same with case 1.
    \item Case 3: A but limit order with price $\$ 93$ and volume 10 arrives. Since there will be no matching counterparty on the ask side, this order will be placed in LOB, resulting in LOB market mid-price and spread shift to $\$ 93.5$ and $\$ 1$.
\end{itemize}

\paragraph{Stylized facts.}
Next, we define stylized facts used in this work following the notation in \citet{coletta2022learning}. We denote $p_a^i(t), v_a^i(t), p_b^i(t), v_b^i(t)$ as the price and volume at $i$-th level of the LOB, at time $t$, for ask and bid respectively. For example, in the starting state of the LOB (left panel) shown in Figure~\ref{fig:LOB_dynamics_illus}, $p_a^1(t) = \$ 94$ and $p_b^1(t) = \$ 92$. The depth of a limit order with price threshold $p(t)$ is defined as
\[d(t)= \begin{cases}p_b^1(t)-p(t), & \text { If side = bid,} \\ p_a^1(t)+p(t), & \text { otherwise.}\end{cases}\]
The stylized facts used here are:
\begin{itemize}
    \item Mid-price, average of ask price and bid price, i.e., \[m(t) = \frac{p_a^1(t) + p_b^1(t)}{2}.\]
    \item Return and price impact: \[r(t_1, t_2) = \log \frac{m(t_1)}{m(t_2)}.\]
    Typically the return at time $t$ is defined as \(r(t, t-1)\); our ``price impact'' is defined as \(r(t, 0)\), which measures the return w.r.t. to the beginning of our market scenario.
    \item Spread: \[\delta(t) = p_a^1(t) - p_b^1(t).\]
    \item Volume imbalance, the demand and supply inequality within the first $n$-levels, i.e.,
    \begin{equation}\label{eq:imbalance}
        I^n(t)=\frac{\sum_{j=1}^n v_b^j(t)}{\sum_{j=1}^n v_b^j(t)+v_a^j(t)}.
    \end{equation}
    \item Absolute volume within the first $n$-levels:
    \[V^n(t)=\sum_{j=1}^n v_b^j(t)+v_a^j(t).\]
    In particular, the bid (or buy) and ask (or sell) volumes within the first $n$-levels are defined as
    \begin{equation}\label{eq:def_vol}
        V^n_b(t)=\sum_{j=1}^n v_b^j(t), \quad V^n_a(t)=\sum_{j=1}^n v_a^j(t).
    \end{equation}
\end{itemize}

\todo{direction feature?}

\subsection{Optimal execution and reinforcement learning}\label{appendix:opt_exec}

One most famous framework for optimal execution (or optimized trade execution), where the goal is to sell a target share of stocks within a finite time horizon, would be the Almgren–Chriss Model \citep{almgren2001optimal}, where the market dynamics takes a parametric form, leading to a ``pre-planned strategy that does not depend on real-time market conditions'' \citep{hambly2021recent}. To address this issue,
\citet{nevmyvaka2006reinforcement} first applied RL in the optimal execution problem. However, their RL training used (historic) real data, in which there will be no responses from other traders.
\citep{karpe2020multi} leveraged the parametric market simulator ABIDES \citep{byrd2020abides,amrouni2021abides} to generate training data to address this issue. On the contrary, \citet{patel2018optimizing} did consider multiple agents in the RL. For recent advancements in this direction, we refer readers to a nice survey by \citet{hambly2021recent} (see Section 4.2 therein).

\paragraph{Reinforcement learning for optimal execution.} 
In the aforementioned optimal execution task, e.g., acquiring $q_0$ shares of one particular asset within $T$, the objective function of finding the optimal policy $\pi^\star$ via online RL problem is:
\[
\pi^\star = \arg\max_\pi \mathbb{E}_\pi\left[\sum_{t=1}^{T} \gamma^{t-1} R(s_{t-1}, a_t) - {\rm P} q_T\right],
\]
where $\gamma \in [0, 1]$ is the discount factor that determines the importance of future rewards, $\mathbb{E}_\pi$ denotes the expectation over trajectories generated by policy $\pi$, $q_T \geq 0$ denotes amount of unfulfilled shares to meet the target $q_0$ at the end of rollout, and ${\rm P} > 0$ is the penalty per unfulfilled share. 

Rule-based policy, such as aggressively placing $q_0$ market orders at the first step, is feasible, but will greatly impact the market and result in sub-optimal profit; Indeed, interacting with the market by executing orders will change the market dynamics, and therefore executing (small) orders sequentially according to the currently observed state is a more sensible strategy. 

\todo{create a table on existing literature: input state for optimal execution}

\paragraph{Deep Q-Network.}
One popular approach to find the optimal policy for optimal execution is to optimize the action-value function, i.e., the Q-function, which represents the expected cumulative discounted reward starting from state $s$, taking action $a$, and following a particular policy. The optimal Q-function, defined as 
$\tilde{Q}^\star(s,a) = \max_\pi \mathbb{E}_\pi [\sum_{k=t}^{T} \gamma^{k-t} R(s_{k-1}, a_k) \,|\, s = s_t, a = a_t]$, obeys the Bellman equation:
\[\tilde{Q}^\star(s, a)=\mathbb{E}_\pi \left[R(s,a)+\gamma \max _{a^{\prime} \in \tilde{\tilde{\mathcal{A}}}} \tilde{Q}^\star \left(s^{\prime}, a^{\prime}\right)  |  s, a\right],\]
where $a^{\prime}$ is the next-state after agent takes action $a$ at state $s$.
When the state and action spaces are discrete, one can estimate the optimal Q-function by iteratively updating the Q-values as follows:
\begin{equation*}
    \begin{split}
        \tilde{Q}_{i + 1}(s, a)  \leftarrow  \tilde{Q}_{i}(s, a) 
        + \alpha \left(R(s, a) + \gamma \max_{a^{\prime}\in\tilde{\mathcal{A}}} \tilde{Q}_{i}(s^{\prime}, a^{\prime}) - \tilde{Q}_{i}(s, a)\right),
    \end{split}
\end{equation*}
where $\alpha$ is the learning rate that controls the weight given to new information. According to \citet{sutton2018reinforcement}, this value iteration algorithm converges to the optimal solution, i.e., $\tilde{Q}_i(s,a) \rightarrow \tilde{Q}^\star(s, a)$ as $i \rightarrow \infty$. In practice, especially when dealing with high-dimensional or continuous state spaces, the Q-function is often approximated using a Deep Q-Network (DQN), i.e., $\tilde{Q}(s, a; \tilde \theta) \approx \tilde{Q}^\star(s, a)$, where a deep NN parameterized by parameter $\tilde \theta$ is used to represent the Q-function, enabling generalization to unseen states \citep{mnih2013playing}. The DQN is trained by minimizing the following objective function (typically via gradient-based method):
\[
\tilde{\cL} (\tilde \theta) = \mathbb{E}\left[\left(R(s, a) + \gamma \max_{a^{\prime}\in\tilde{\mathcal{A}}} \tilde{Q}(s^{\prime}, a^{\prime}; \tilde \theta) - \tilde{Q}(s, a; \tilde \theta)\right)^2\right].
\]

\subsection{Baseline Simulator: Conditional Wasserstein GAN}\label{appendix:GAN}
Despite the success of variational autoencoder (VAE) \citep{sohn2015learning} under many other contexts, the most popular machine learning approach for generating market data is still GAN.
Conditional GAN (cGAN) \citep{mirza2014conditional} extends the standard GAN \citep{goodfellow2014generative} framework by incorporating conditional information. The generator network, which acts like a trading agent, \(G_\theta\) takes random noise input \(z\) and a conditional variable \(S\) to produce synthetic orders $A$ (i.e., random sample $z$ is generated from a prior distribution, such as Gaussian distribution, and then transformed by $G_\theta$ to ``match'' the true distribution).
The discriminator network \(D\) estimates the probability that an order is real by considering both the generated order \(G_\theta(z|S)\) and the true order \(A\) along with the corresponding conditional variable \(S\). The training of a cGAN involves optimizing a min-max objective function:
\begin{equation*}
\begin{split}
    \min_\theta \max_D \  \mathbb{E}_{A|S \sim p_{\real}}[\log D(A|S)]  + \mathbb{E}_{z \sim p_{\rm z}}[\log(1 - D(G_\theta(z|S)))],
\end{split}  
\end{equation*}
where \(p_{\real}\) represents the real data distribution and \(p_{\rm z}\) represents the prior noise distribution, and $S$ represents the feature/input that cGAN will condition on.

Even though there have been modern machine learning techniques adapted to generative models under various specific domains, such as imitation learning for traffic simulation to test self-driving algorithms \citep{suo2021trafficsim}, such adaption for LOB stock market simulation is largely missing until \citet{li2020generating,coletta2021towards,coletta2022learning}, who leveraged cWGANs to train a NN-based world BG agent that places realistic orders based on relevant market information. By utilizing a cWGAN, the world BG agent can learn from the simulated market data, improving its trading strategies in response to the dynamic market. In particular, \citet{coletta2021towards} used Wasserstein distance as the discriminator, i.e., they adopted
Wasserstein GAN \citep{arjovsky2017wasserstein}: 
\[ \min_\theta \max _{w \in \mathcal{W}} \mathbb{E}_{A|S \sim p_\real}\left[f_w(A|S)\right]-\mathbb{E}_{z \sim p_{\rm z}}\left[f_w\left(G_\theta(z|S)\right)\right],\]
where $f_w$ is the discriminator and the function $G_\theta$ is the generator. They used gradient descent w.r.t. $w$ and $\theta$ iteratively to optimize the objective. Additionally, there exist methods to calibrate ABIDES with advanced machine learning techniques such as GAN \citep{storchan2021learning,shi2023neural}, the mainstream market simulator is still based on GAN due to its superior performance; another main reason is that the agent identity information is missing is real market data, making it difficult to calibrate the parametric market simulator.

\newpage

\section{Additional Experimental Details of Algorithmic Trading-guided Market Simulation}
\subsection{Training details}\label{appendix:training}

\paragraph{Configuration.}
In our experiment, the configuration of the optimal execution task is: time horizon $T = 10$, parent order size $q_0 = 50$.
For the AT agent performing this task, the private state at time step $t$ for the AT agent is normalized elapsed time $t/T$, remaining shares to acquire normalized by total parent order size $q_0$, and their difference; The market state consists of imbalance (\eqref{eq:imbalance} with all levels and $n = 5$), spread, price impact, and direction feature.
The action space $\tilde{\mathcal{A}} = \{0,1,2\}$, where $0$ stands for executing fixed size Market Order,
$1$ stands for executing fixed size ($10$ shares) Limit Order, and
$2$ stands for action hold (i.e., no action). 
In our experiments, the external agent to places orders aggressively such that the world BG agent is able to encounter diverse market states. Additionally, this eases the implementation burden as rollout using this simple rule-based AT agent (instead of NN-based agent) can be performed in parallel easily, which leads to reduced computational cost. In particular, we choose an aggressive agent which continues to place limit orders until the parent order is fulfilled, which results in $T=5$ in practice. The parametric BG agents' (treated as ``reality'' in the experiment) configuration is: one Exchange Agent, two Adaptive Market Maker Agents, $100$ Value Agents, $25$ Momentum Agents, and $5000$ Noise Agents; Please see \citet{byrd2020abides,amrouni2021abides} for additional details on the parametric market simulator ABIDES\footnote{Available at \url{https://github.com/jpmorganchase/abides-jpmc-public}}.


\paragraph{Training hyperparameters.}
As mentioned previously, the complete rollout number is chosen as $N = 5$ such that the feedback's empirical distribution under the world BG agent market is sufficiently different from that of the ``real market''. To ensure rollouts under the ``real market'' will not incur extreme feedbacks, we conduct $N' = 100$ offline rollouts to construct the feedback collection under world market. As $T_0$ and $b$ together act like the batch-size, we consider grid search over $T_0 \in \{3,4,5\}$ and $b \in \{3,5\}$. By this grid search, we choose the combination with the smallest (proposed) metric, which is
\begin{itemize}
    \item $T_0 = 5$ and $b = 3$ for \texttt{Mkt2NextReturn}, which corresponds to the results in Figure~\ref{fig:exp_effectiveness_1} and the \textbf{second} column in Figures~\ref{fig:exp_comparison_1}, \ref{fig:exp_comparison_more_volume}, \ref{fig:exp_comparison_others_same} and \ref{fig:exp_comparison_others_diff}.
    \item $T_0 = 5$ and $b = 3$ for \texttt{Mkt2NextPriceImpact}, which corresponds to the results in the \textbf{forth} column in Figures~\ref{fig:exp_comparison_1}, \ref{fig:exp_comparison_more_volume}, \ref{fig:exp_comparison_others_same} and \ref{fig:exp_comparison_others_diff}.
\end{itemize}
To compare the results of \texttt{Mkt2NextReturn} with \texttt{EpisodeReward}, we use the same hyperparameter setting for \texttt{EpisodeReward}, i.e.,
\begin{itemize}
    \item $T_0 = 5$ and $b = 3$ for \texttt{EpisodeReward}, which corresponds to the results in the \textbf{third} column in Figures~\ref{fig:exp_comparison_1}, \ref{fig:exp_comparison_more_volume}, \ref{fig:exp_comparison_others_same} and \ref{fig:exp_comparison_others_diff}. The corresponding AT-based metric is optimized from 0.7156 to 0.6176.
\end{itemize}
Additionally, to further demonstrate of the effectiveness of our AT-based metric and ATMS, we consider
\begin{itemize}
    \item $T_0 = 2$ and $b = 5$ for \texttt{Mkt2NextReturn}, which corresponds to the results in Figure~\ref{fig:exp_effectiveness_2}.
\end{itemize}
The initial learning rate $r$ is chosen to be $10^{-9}$, which decreases by half every $10$ iteration. There are in total $100$ iterations and based on the decrease of learning rate they are divided into $10$ epochs (each with $10$ iterations).
We use the cWGAN network parameter trained on AINV market reply data (which is different from the ``real'' market) as the warm-start/initialization when we train ATMS.

\subsection{Additional results for other feedback candidates}\label{appendix:f}

\begin{figure}[!htp]
\centerline{
\includegraphics[width = .65\textwidth]{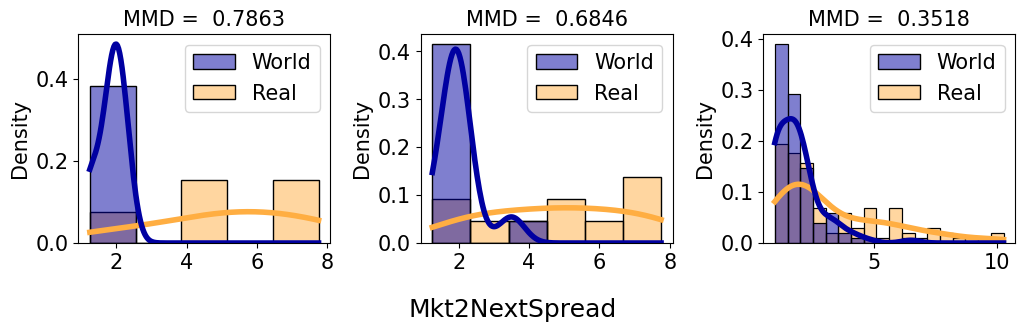}
}
\centerline{
\includegraphics[width = .65\textwidth]{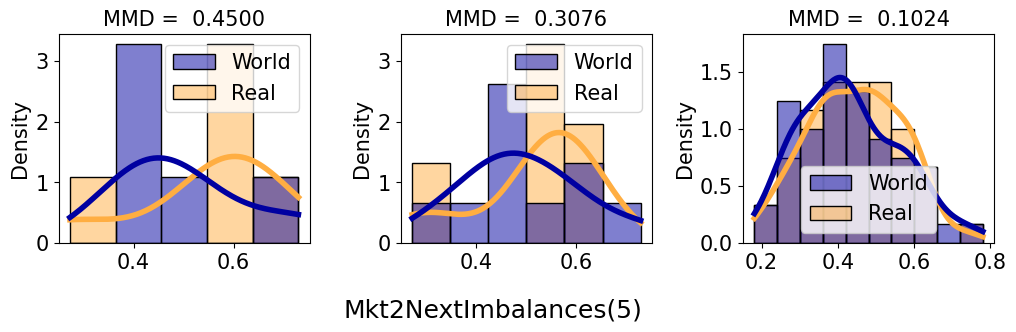}
}
\centerline{
\includegraphics[width = .65\textwidth]{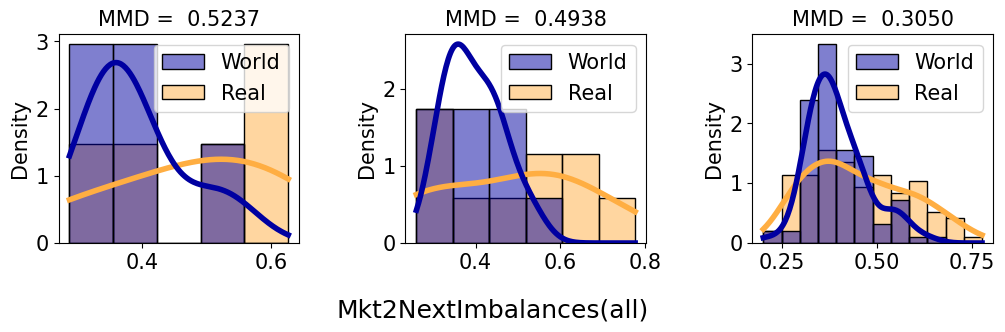}
}
\centerline{
\includegraphics[width = .65\textwidth]{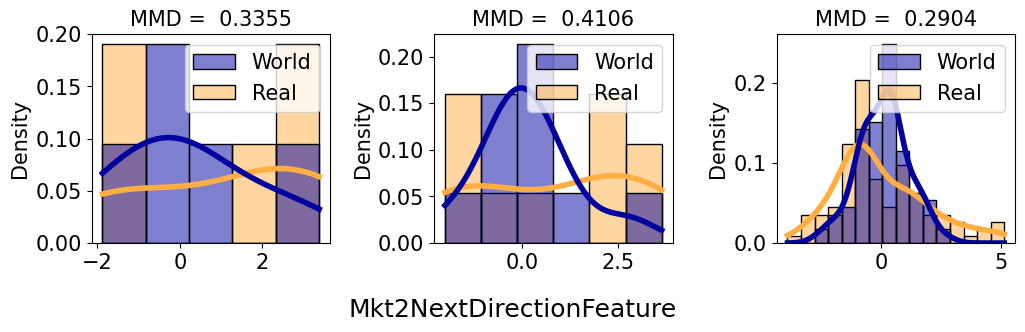}
}
\centerline{
\includegraphics[width = .65\textwidth]{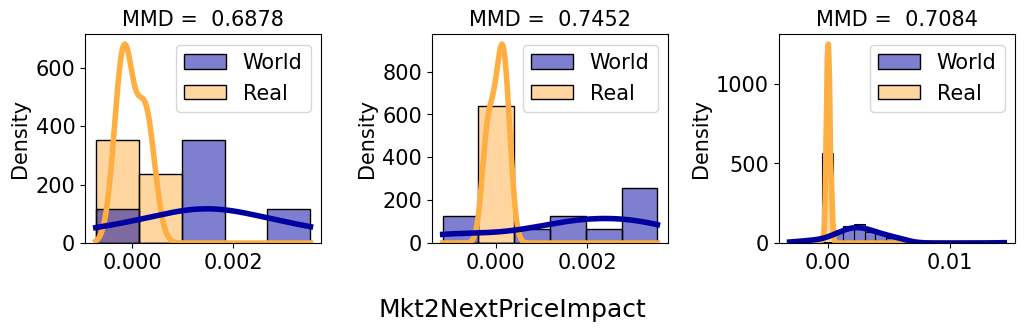}
}
\caption{The empirical distribution of additional causal effect feedback candidates (from top to bottom, we report \texttt{Mkt2Next\underline{Spread}}, \texttt{Mkt2Next\underline{Imbalances(5)}}, \texttt{Mkt2Next\underline{Imbalances(All)}}, 
\texttt{Mkt2Next\underline{Direction}}, and 
\texttt{Mkt2Next\underline{PriceImpact}}) when performing multiple (from left to right: $5, 10, 100$) rollouts using AT agent under real (orange) or world BG agent (blue) markets. Again, the idea feedback should be able to differentiate different markets with only a few rollouts, making the \texttt{Mkt2NextPriceImpact} the most ideal candidate in this figure.}\label{fig:exp2-4}
\end{figure}

In Figure~\ref{fig:exp2-4}, we present the empirical distribution of other feedback candidates. Interestingly, when utilizing various stylized facts, with the exception of price impact, as the next state, we observe a similar pattern to \texttt{EpisodeReward}. This similarity renders these alternatives unfavorable choices in our proposed ATMS and might suggest that those stylized facts are less representative of the market. It is important to mention that we observe a similar pattern of the results for \texttt{Mkt2NextPriceImpact} (shown in the last row of Figure~\ref{fig:exp2-4}) to that of our ideal candidate \texttt{Mkt2NextReturn} (shown in the second row of Figure~\ref{fig:exp2-1} in the main text of the paper). Therefore, \texttt{Mkt2NextPriceImpact} is another candidate that can be readily used to train/improve the world BG agent.


\subsection{Additional results for other $\hat d$ candidates}\label{appendix:hatd}  
\begin{figure}[!htp]
\centerline{
\includegraphics[width = .355\textwidth]{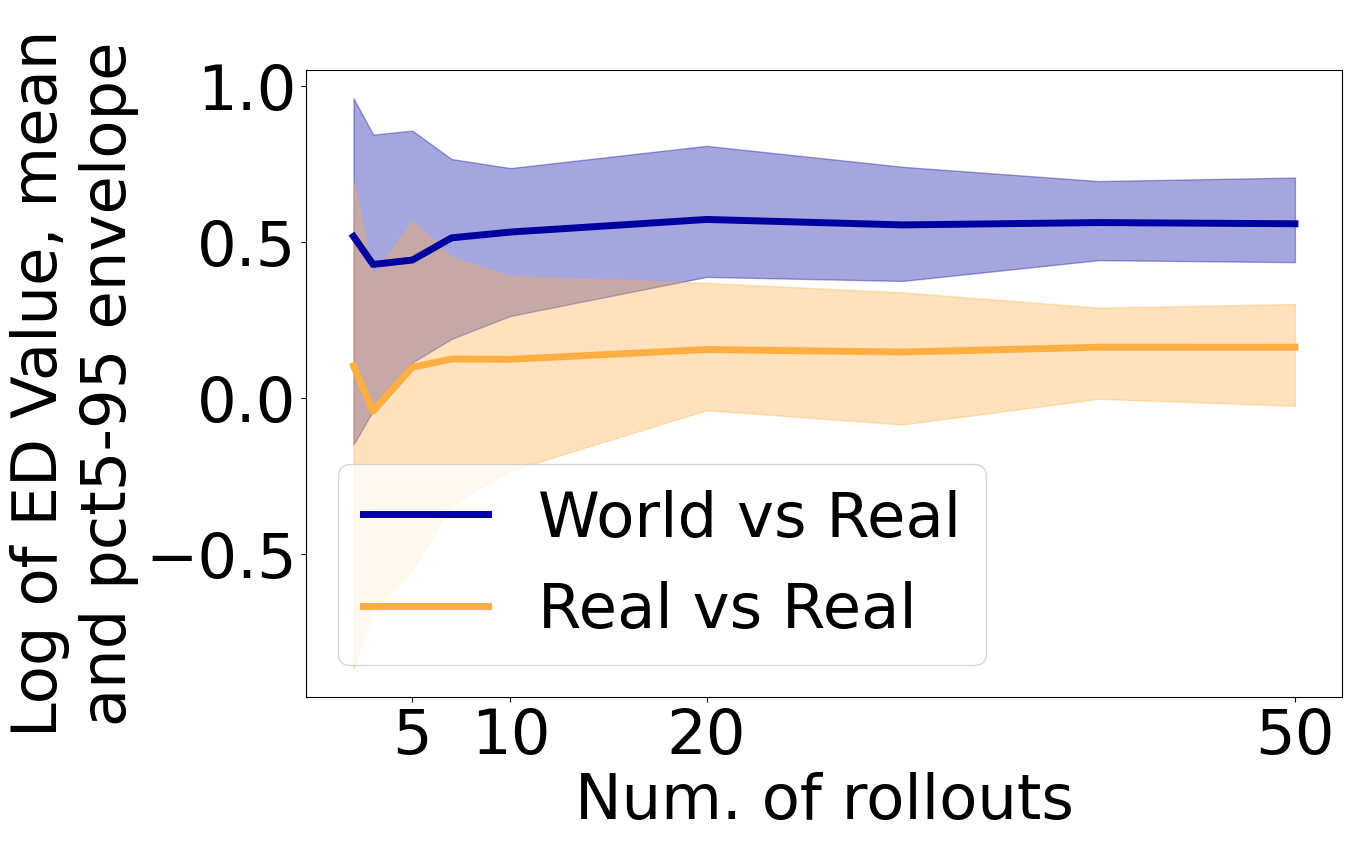}
\hspace{0.1in}
\includegraphics[width = .33\textwidth]{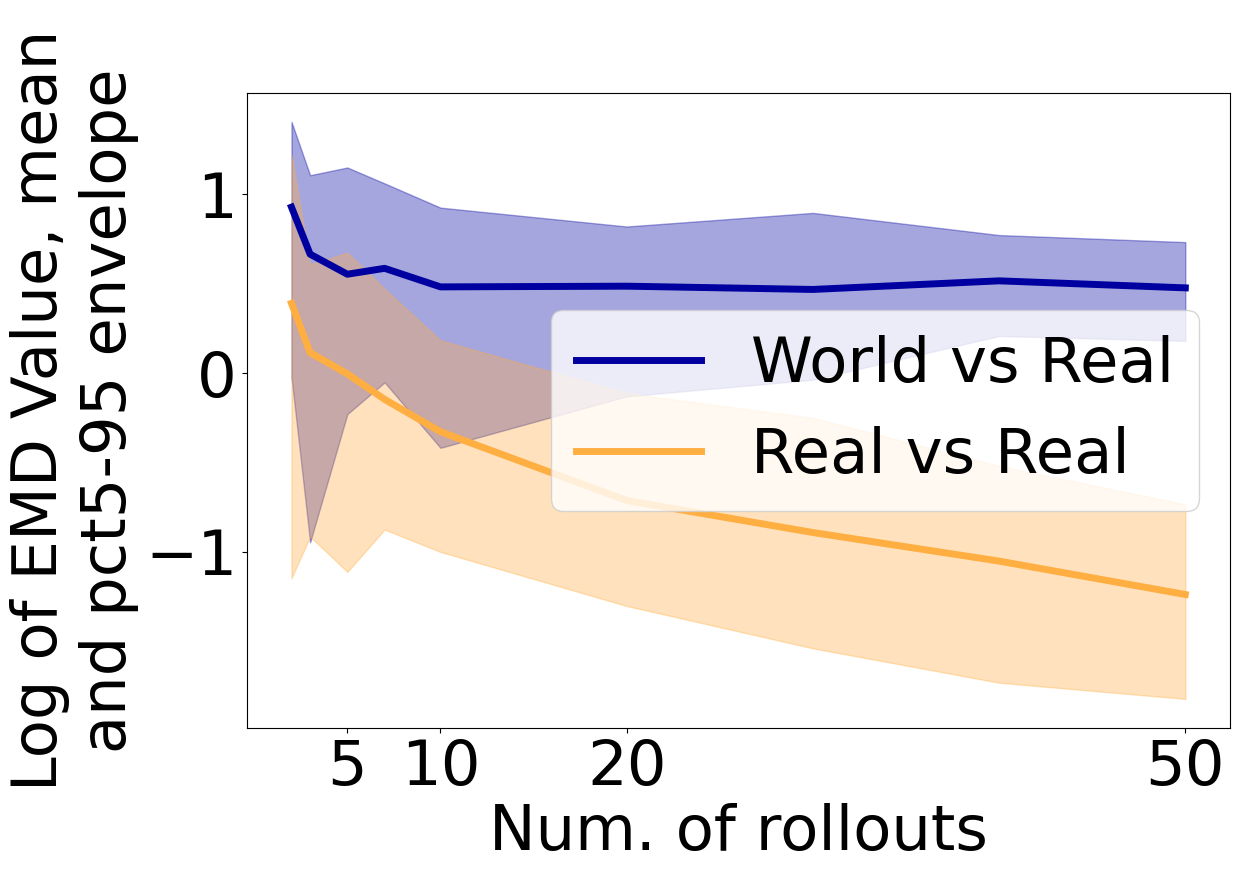}
}
\centerline{
\includegraphics[width = .345\textwidth]{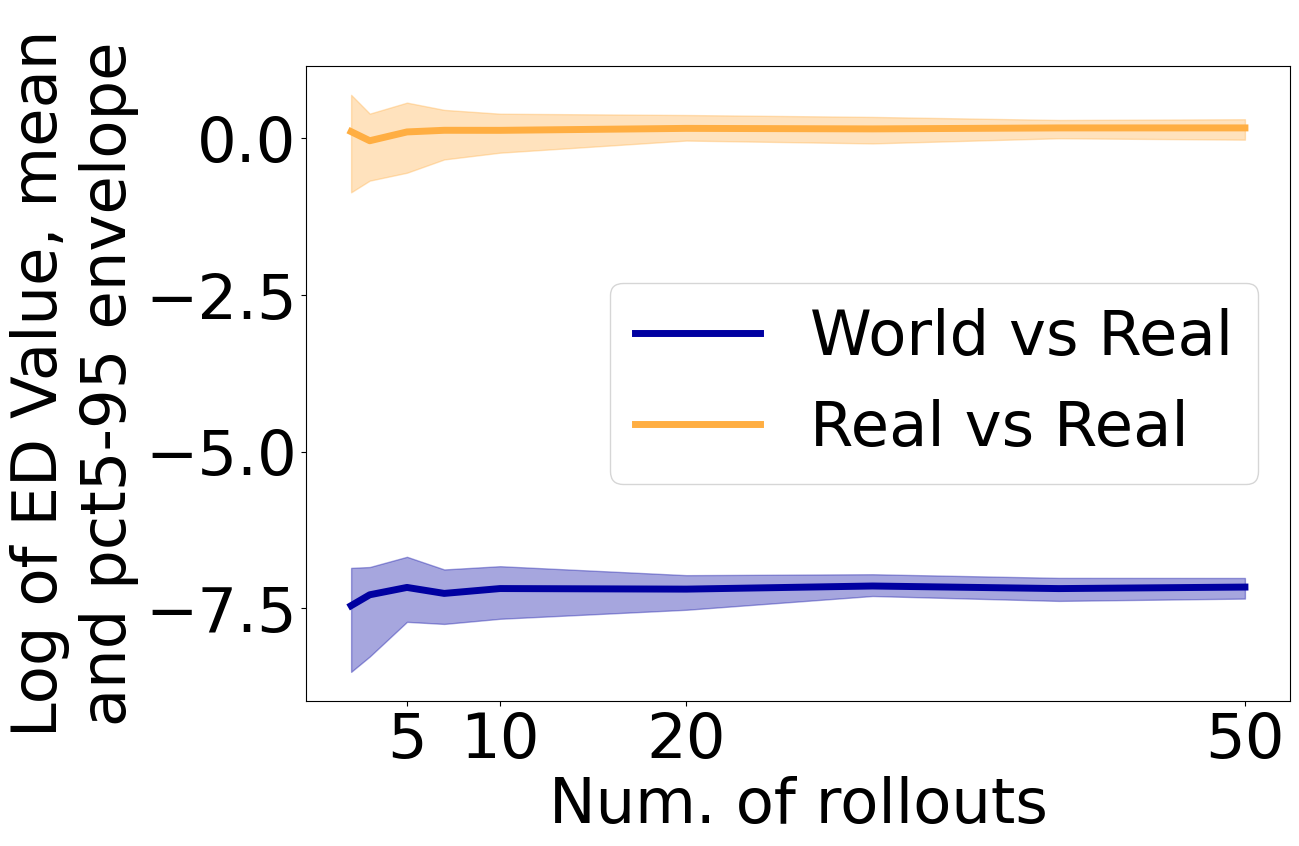}
\hspace{0.1in}
\includegraphics[width = .34\textwidth]{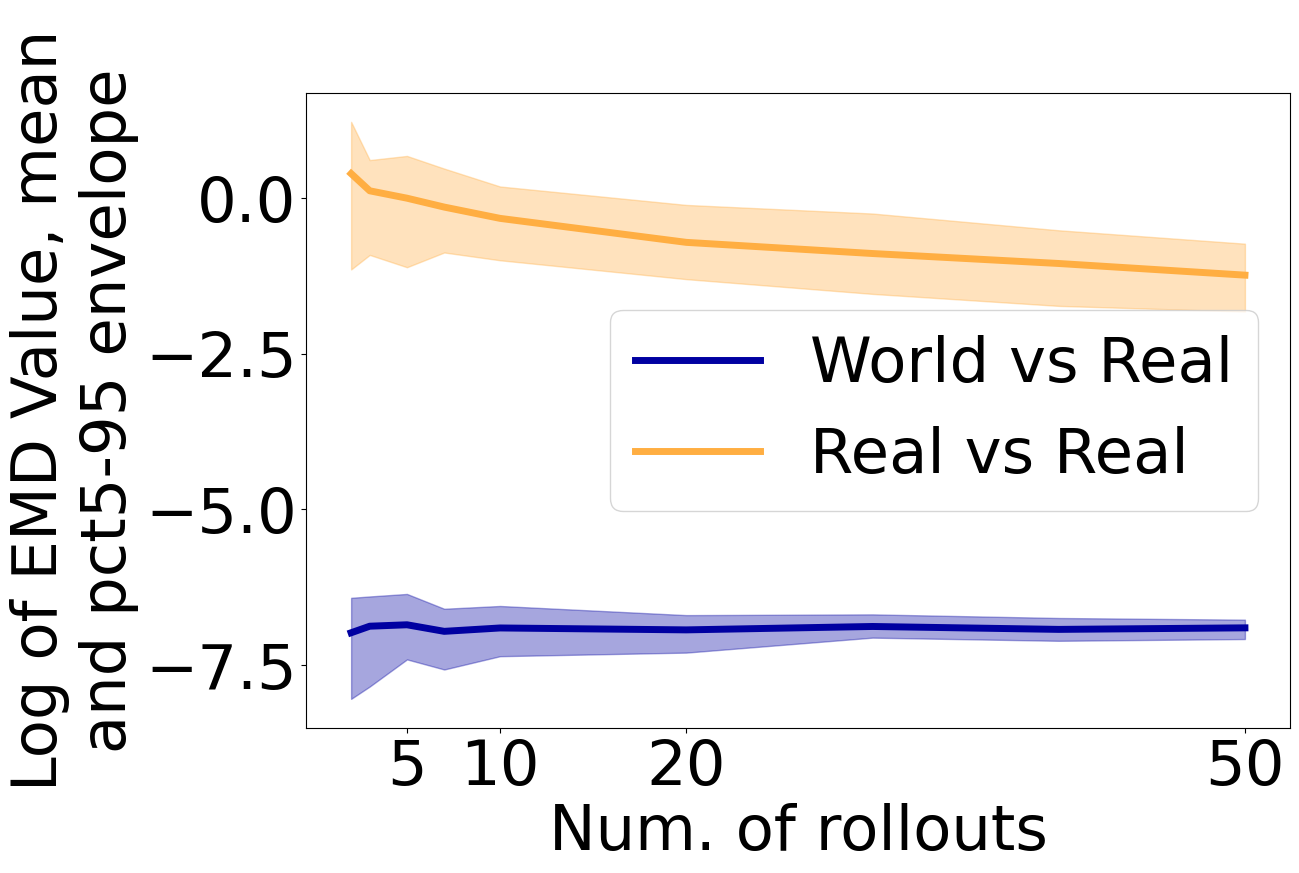}
}
\caption{The mean and $5 - 95 \%$ envelop trajectory (over 50 bootstrap trials) against increasing number of rollouts of our proposed market distance metric with feedbacks \texttt{EpisodeReward} (top) and \texttt{Mkt2NextReturn} (bottom), and $\hat d$ chosen to be ED (left) and EMD (right). The above results all indicate poor $f$ and $\hat d$ combinations that cannot be used for INTAGS (or ATMS) training.}
\label{fig:exp3-2} 
\end{figure}

In addition to the result for MMD depicted in Figure~\ref{fig:exp3-1}, we present the outcomes for Energy Distance and Earth Mover's Distance in Figure~\ref{fig:exp3-2}. In the first row where $f$ is chosen as \texttt{EpisodeReward}, we can see the resulting market distance metric's behavior is reasonable in the sense that the metric is closer to zero when comparing two identical markets (indicated by the orange line, ``Real vs Real''). However, the metric performs poorly to differentiate markets as we cannot see the pattern that blue line is significantly larger than the orange one with small rollout number (as shown in right panel in Figure~\ref{fig:exp3-1}).

From the second row in Figure~\ref{fig:exp3-2}, where \texttt{Mkt2NextReturn} is chosen as $f$,
it is surprising to find that the metric's behavior appears unreasonable as the metric between different markets is smaller than that when two markets are identical. The above observations indicate that the corresponding metrics may not be suitable for capturing the dissimilarity (or similarity) between different markets accurately, and therefore ED and EMD are less favorable candidates for the subsequent simulator training.

\begin{figure}[!htp]
\centerline{
\includegraphics[width = .33\textwidth]{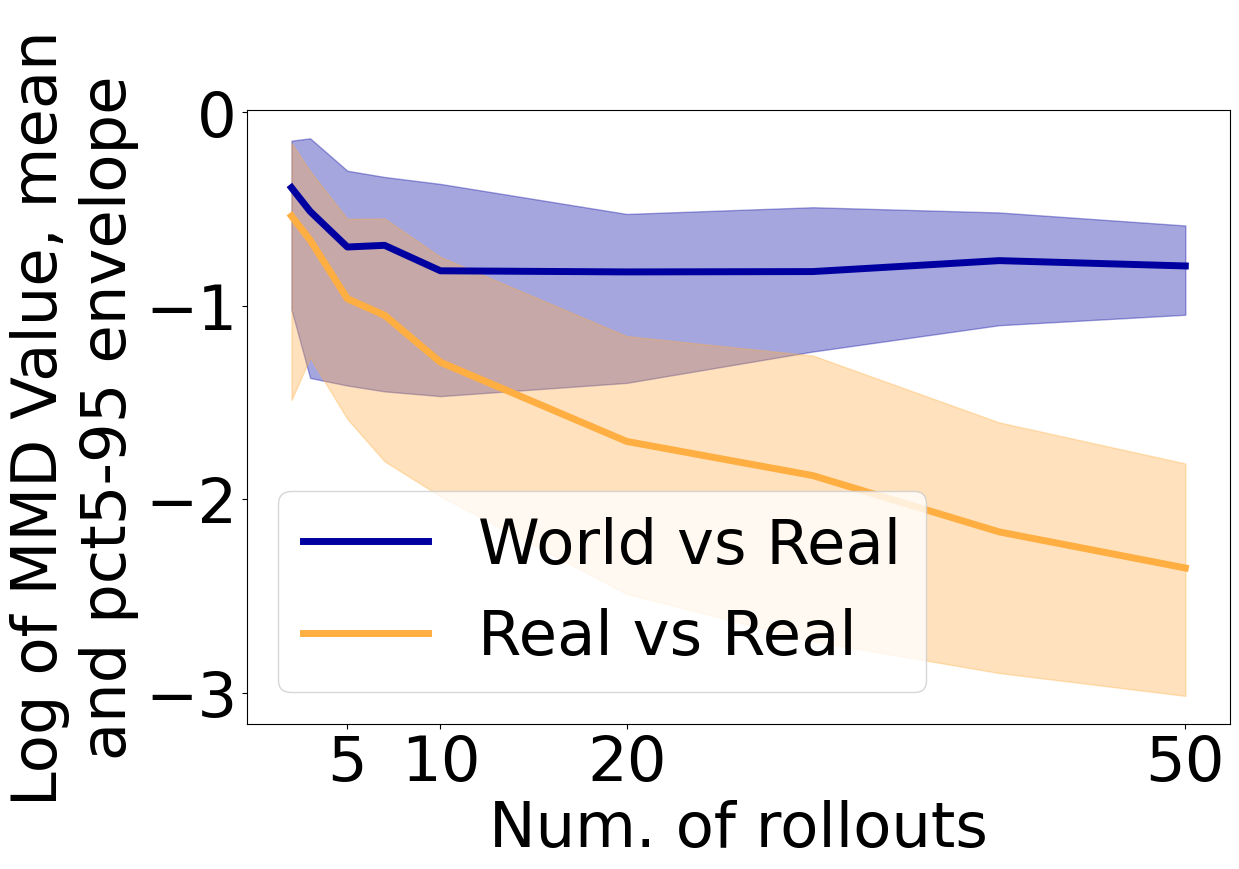}
\includegraphics[width = .33\textwidth]{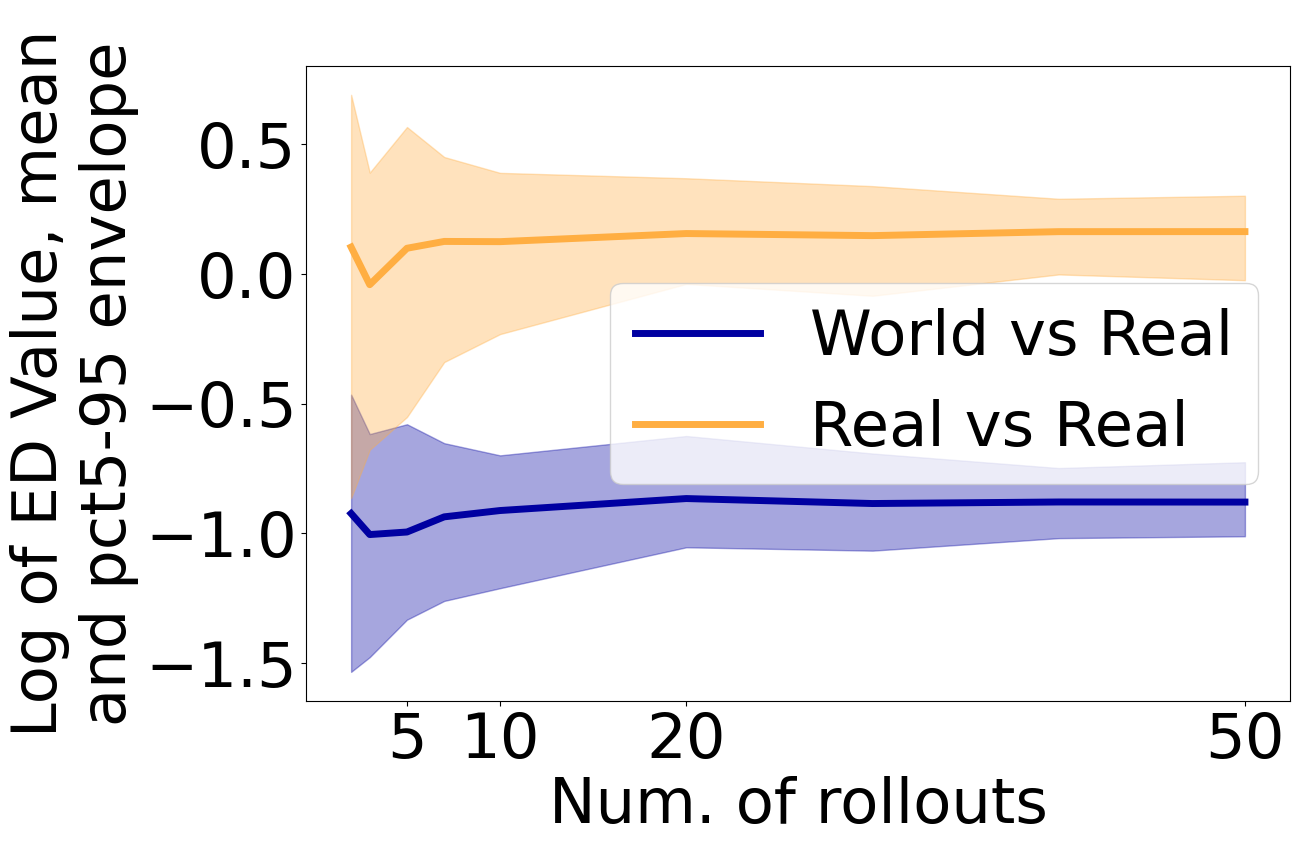}
\includegraphics[width = .33\textwidth]{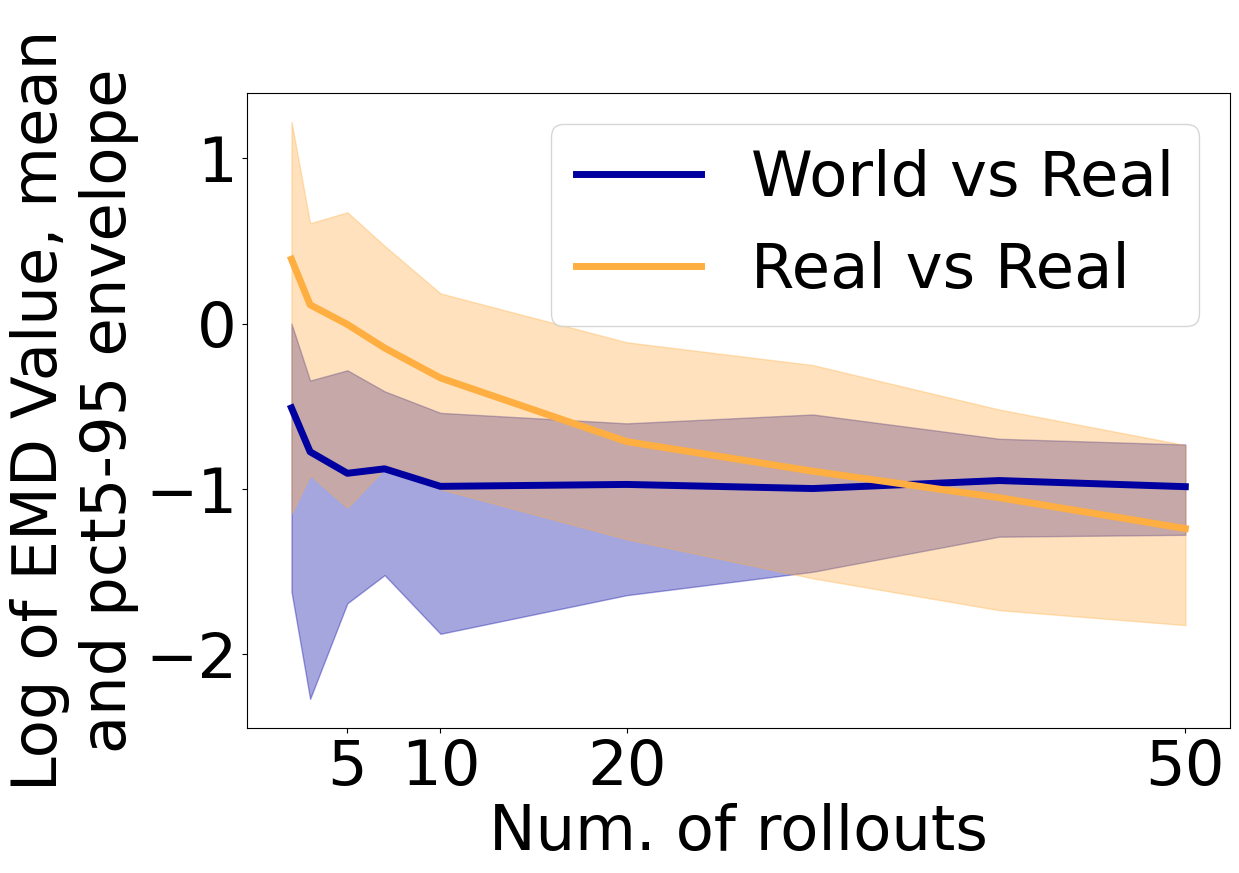}
}
\caption{The mean and $5 - 95 \%$ envelop trajectory (over 50 bootstrap trials) against increasing number of rollouts of our proposed market distance metric with feedback \texttt{Mkt2Reward}, and $\hat d$ chosen to be MMD (left), ED (middle) and EMD (right). Similarly, the above results all indicate poor $f$ and $\hat d$ combinations that cannot be used for INTAGS (or ATMS) training.}
\label{fig:exp3-3} 
\end{figure}

For completeness and to provide further insights into their unsuitability (for subsequent training of ATMS), we also include the results for MMD, ED, and EMD when choosing \texttt{Mkt2Reward} as feedback \(f\) in Figure~\ref{fig:exp3-3}. These additional results serve as direct evidence highlighting the undesirable characteristics of these metrics in the context of our ATMS. The findings reinforce the significance of developing and employing a specialized distance metric, as we propose, to effectively evaluate the distance between financial markets, ensuring the reliability of our ATMS.

\subsection{Additional evidence for the effectiveness of ATMS}\label{appendix:effectiveness} 
\begin{figure}[!htp]
\centerline{
\includegraphics[width = \textwidth]{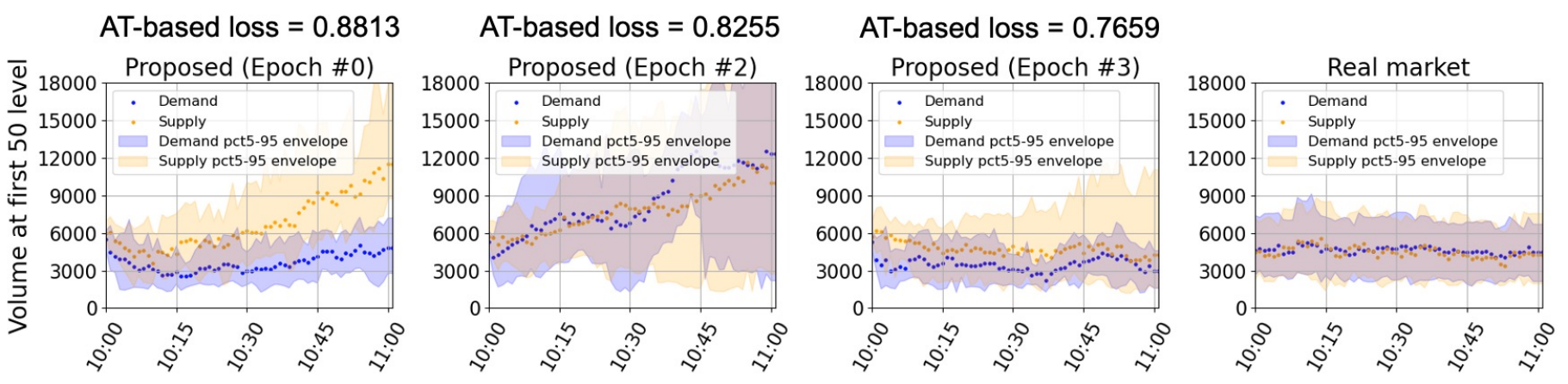}
}
\caption{Effectiveness evaluation of our proposed ATMS with different random seeds and training hyperparameters. The consistent pattern across experiments reaffirms the robustness of ATMS in generating realistic market data.}
\label{fig:exp_effectiveness_2}
\end{figure}

To further establish the reliability of our proposed ATMS, we conduct additional experiments to investigate the impact of randomness and training hyperparameters on its effectiveness. In Figure~\ref{fig:exp_effectiveness_2}, we present the results obtained using a different random seed and distinct training hyperparameters. Remarkably, we observe a similar pattern to the findings depicted in Figure~\ref{fig:exp_effectiveness_1}. This consistency across different experimental setups strongly suggests that the effectiveness demonstrated in Figure~\ref{fig:exp_effectiveness_1} is not solely attributed to randomness but indeed indicative of the superior performance of our proposed ATMS. These findings reinforce the credibility and practical applicability of our ATMS, affirming its capability to consistently generate realistic market data regardless of varying random initialization and training conditions.

\subsection{Additional comparison results for \SFVn}\label{appendix:otherSF} 
\begin{figure*}[!htp]
\centerline{
\includegraphics[trim={0.25cm 0.25cm 0.25cm 0.25cm},clip,width = .2065\textwidth]{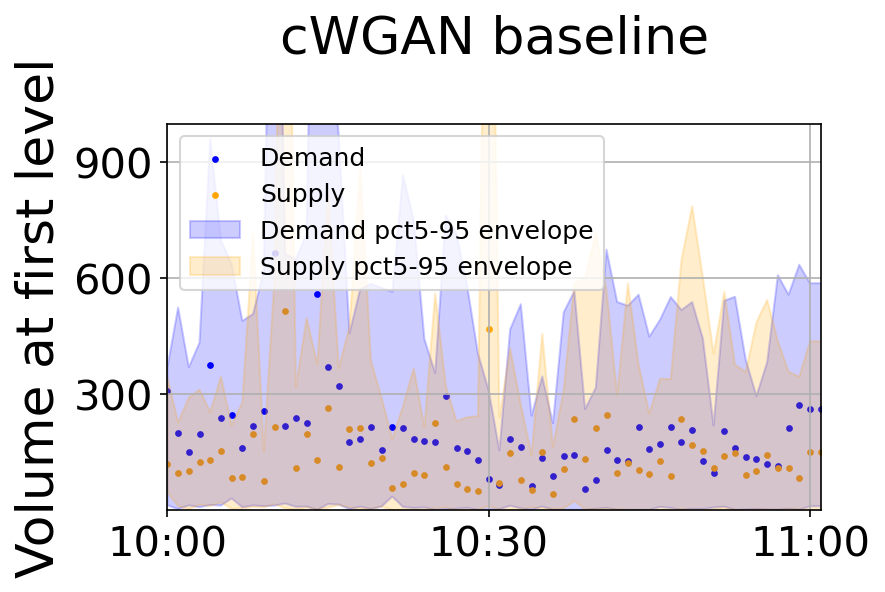}
\includegraphics[trim={1.3cm 0.25cm 0.25cm 0.25cm},clip,width = .19\textwidth]{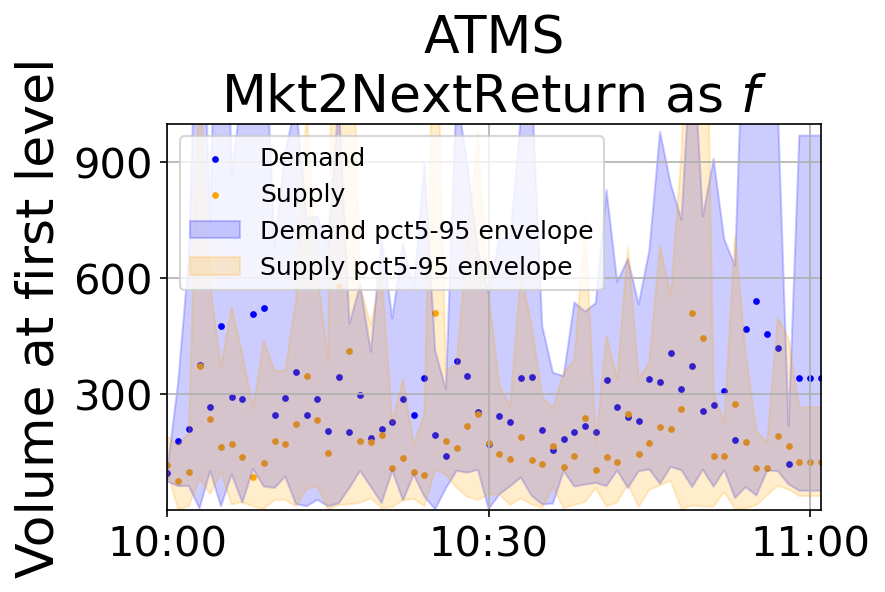}
\includegraphics[trim={1.3cm 0.25cm 0.25cm 0.25cm},clip,width = .19\textwidth]{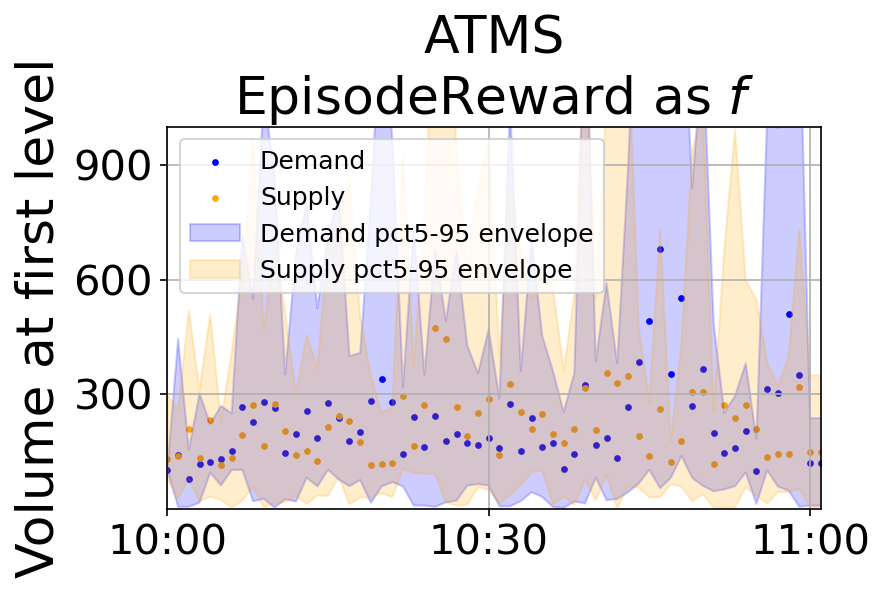}
\includegraphics[trim={1.3cm 0.25cm 0.25cm 0.25cm},clip,width = .19\textwidth]{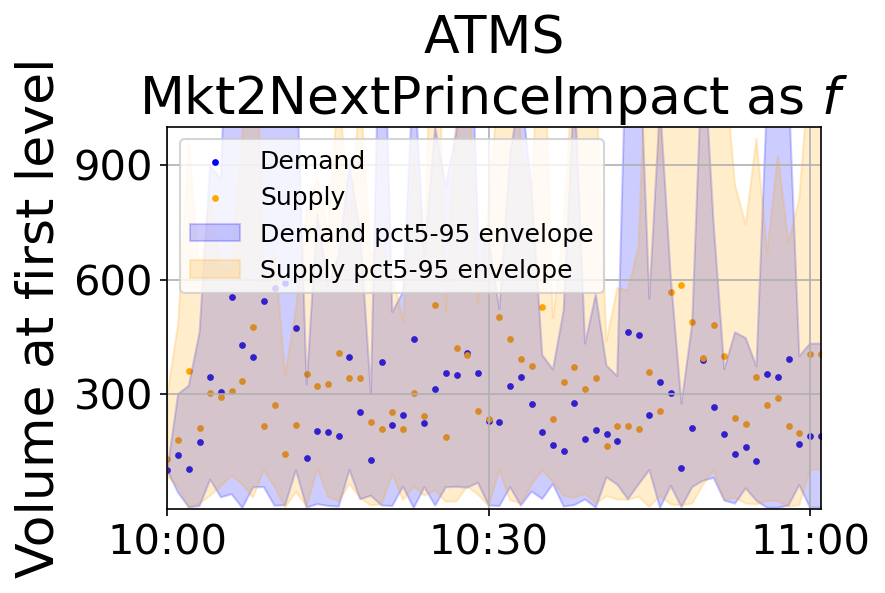}
\includegraphics[trim={1.3cm 0.25cm 0.25cm 0.25cm},clip,width = .19\textwidth]{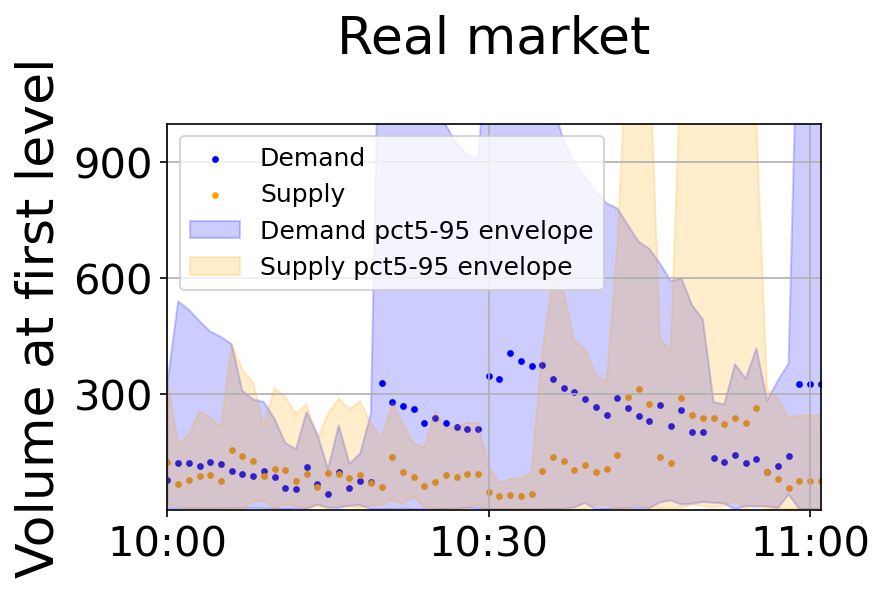}
}
\centerline{
\includegraphics[trim={0.25cm 0.25cm 0.25cm 0.25cm},clip,width = .2065\textwidth]{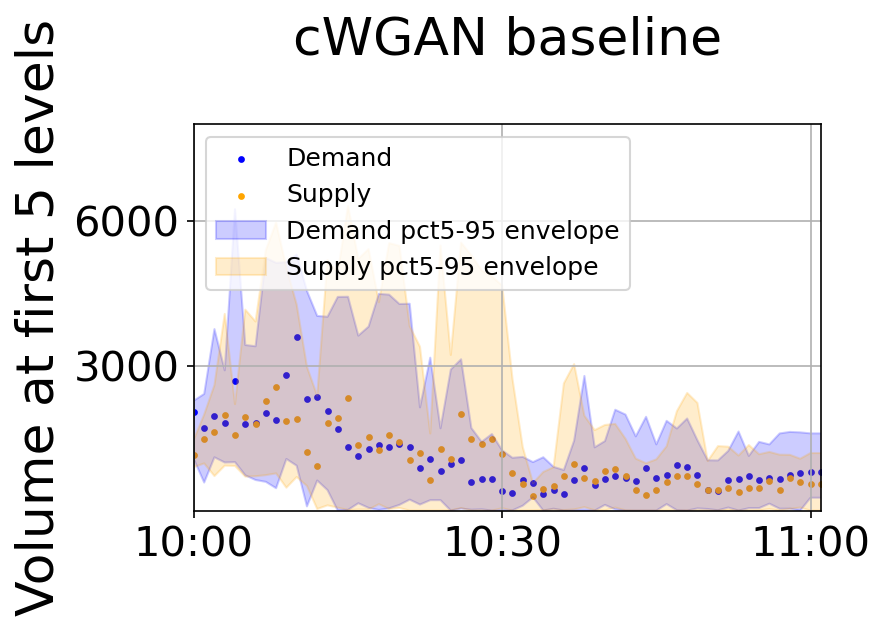}
\includegraphics[trim={1.3cm 0.25cm 0.25cm 0.25cm},clip,width = .19\textwidth]{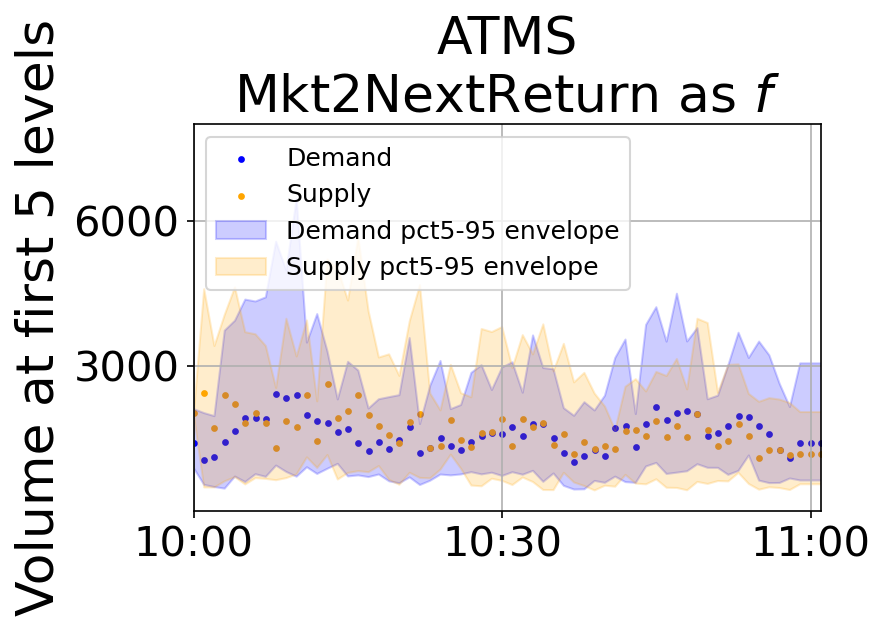}
\includegraphics[trim={1.3cm 0.25cm 0.25cm 0.25cm},clip,width = .19\textwidth]{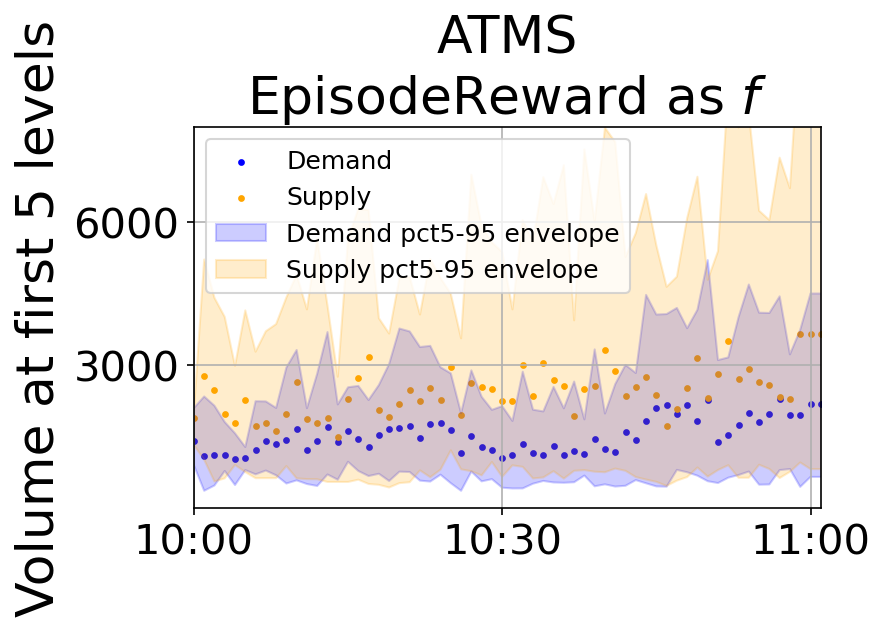}
\includegraphics[trim={1.3cm 0.25cm 0.25cm 0.25cm},clip,width = .19\textwidth]{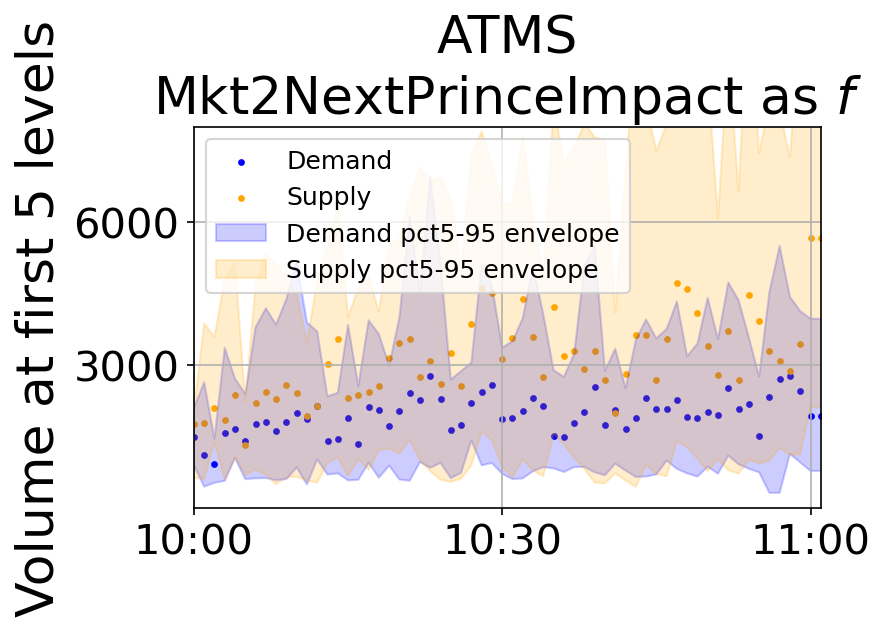}
\includegraphics[trim={1.3cm 0.25cm 0.25cm 0.25cm},clip,width = .19\textwidth]{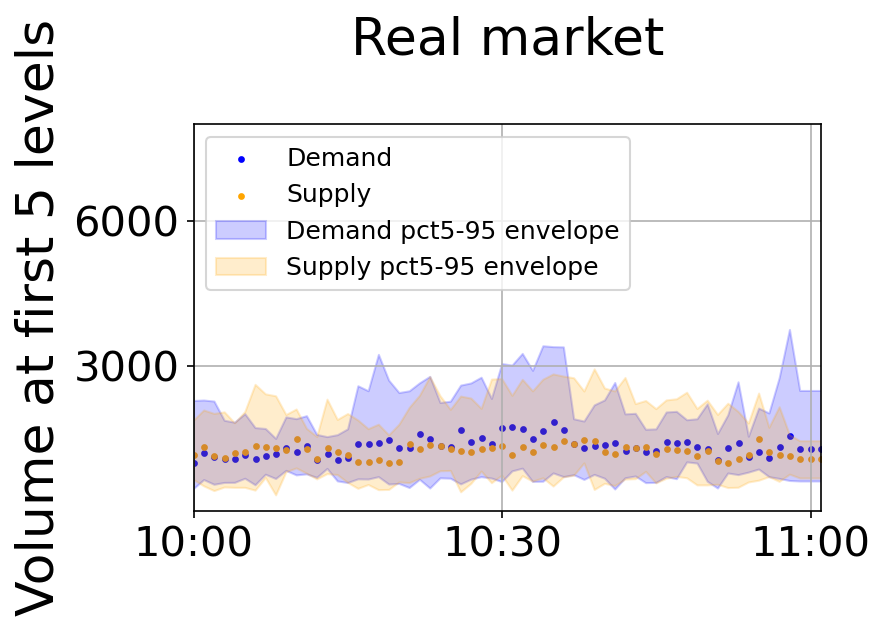}
}
\caption{Comparison of \SFVn \ for $n \in \{1,5\}$ among different simulated markets (specified on top of each panel). We can observe that neither our ATMS nor the cWGAN baseline can correctly capture the volume at the first level whereas our ATMS with \texttt{Mkt2NextReturn} can correctly capture the volume at the first $5$-levels (just as shown in Figure~\ref{fig:exp_comparison_1}).}
\label{fig:exp_comparison_more_volume}
\end{figure*}

In the last part, we report results of \SFVn \ for $n \in \{1,5\}$ in Figure~\ref{fig:exp_comparison_more_volume}, from which we can observe: our proposed approach is behaving similarly to cWGAN baseline for $n=1$ case, and they both do not fully capture the real market dynamics; For $n=5$ case, the patterns are fairly similar to that of $n = 10$ case shown in Figure~\ref{fig:exp_comparison_1}, reaffirming our claims above.

\end{document}